\documentclass[twoside]{article}
\usepackage{palatino,epsfig,latexsym,natbib}
\usepackage{xspace}
\newtheorem{lemma}{Lemma}

\newtheorem{theorem}{Theorem}
\newtheorem{proposition}{Proposition}

\newtheorem{corollary}{Corollary}
\newcommand{\ie}{i.\,e. \ }
\newcommand{\unimodal}{{\sc Unimodal}\xspace} 
\newcommand{\vcp}{{VCP}\xspace}
\newcommand{\scp}{{SCP}\xspace}
\newcommand{\twosat}{{2-SAT}\xspace}
\newcommand{\onemax}{{\sc OneMax}\xspace}
\newcommand{\oneplusoneEA}{($1$+$1$)~EA\xspace}
\newcommand{\onelambdaEA}{($1$,$\lambda$)~EA\xspace}
\newcommand{\oneoneEA}{($1$,$1$)~EA\xspace}
\newcommand {\A}{\mbox{\bf A}}
\newcommand {\B}{\mbox{\bf B}}
\newcommand {\PP}{\mbox{\bf P}}
\newcommand {\Q}{\mbox{\bf Q}}
\newcommand {\T}{\mbox{\bf T}}
\newcommand {\E}{\mbox{\bf E}}
\newcommand {\x}{\mbox{\bf x}}
\newcommand {\p}{\mbox{\bf p}}
\newcommand {\uu}{\mbox{\bf u}}
\newcommand {\y}{\mbox{\bf y}}
\newcommand {\z}{\mbox{\bf z}}

\newcommand {\W}{{\bf W}}

\newcommand {\K}{{\mathcal K}}
\def\R{\mathsf{l\kern -.15em R}}

\newcommand {\G}{{\bf \Gamma}}

\begin{document}

\title{On Proportions of Fit Individuals in Population of Mutation-Based
Evolutionary Algorithm with Tournament Selection}

\author{Anton V. Eremeev\thanks{$^*$ Supported by Russian
Foundation for Basic Research grants 15-01-00785 and
16-01-00740.}}

\maketitle

\sloppy

\begin{center}
Sobolev Institute of Mathematics, Omsk Branch,\\
        13 Pevtsov str., Omsk, 644099, Russia\\
        eremeev@ofim.oscsbras.ru\\
        \end{center}

\begin{abstract}

In this paper, we consider a fitness-level model of a non-elitist
mutation-only evolutionary algorithm~(EA) with tournament
selection. The model provides upper and lower bounds for the
expected proportion of the individuals with fitness above given
thresholds. In the case of so-called monotone mutation, the
obtained bounds imply that increasing the tournament size improves
the EA performance. As corollaries, we obtain an exponentially
vanishing tail bound for the Randomized Local Search on unimodal
functions and polynomial upper bounds on the runtime of EAs on
2-SAT problem and on a family of Set Cover problems proposed by
E.~Balas.
\end{abstract}


\section{Introduction \label{intro}}

Evolutionary algorithms are randomized heuristic algorithms
employing a population of tentative solutions (individuals) and
simulating an evolutionary type of search for optimal or
near-optimal solutions by means of selection, crossover and
mutation operators. The evolutionary algorithms with crossover
operator are usually called genetic algorithms~(GAs). Evolutionary
algorithms in general have a more flexible outline and include
genetic programming, evolution strategies, estimation of
distribution algorithms and other evolution-inspired paradigms.
Evolutionary algorithms are now frequently used in areas of
operations research, engineering and artificial intelligence.

Two major outlines of an evolutionary algorithm are the {\em
elitist} evolutionary algorithm, that keeps a certain number of
most promising individuals from the previous iteration, and the
{\em non-elitist} evolutionary algorithm, that computes all
individuals of a new population independently using the same
randomized procedure. In this paper, we focus on the non-elitist
case.

One of the first theoretical results in the analysis of
non-elitist GAs is Schemata Theorem~\citep{bib:Gold89} which gives
a lower bound on the expected number of individuals from some
subsets of the search space (schemata) in the next generation,
given the current population. A significant progress in
understanding the dynamics of GAs with non-elitist outline was
made in~\citep{Vo} by means of dynamical systems. However most of
the findings in~\citep{Vo} apply to the infinite population case,
and it is not clear how these results can be used to estimate the
applicability of GAs to practical optimization problems. A
theoretical possibility of constructing GAs that provably optimize
an objective function with high probability in polynomial time was
shown in~\citep{V00} using rapidly mixing Markov chains.
However~\citep{V00} provides only a very simple artificial example
where this approach is applicable and further developments in this
direction are not known to us.

One of the standard approaches to studying evolutionary algorithms
in general, is based on the {\em fitness
levels}~\citep{bib:Wegener2002}. In this approach, the solution
space is partitioned into disjoint subsets, called fitness-levels,
according to values of the fitness function. In~\citep{bib:l11},
the fitness-level approach was first applied to upper-bound the
runtime of non-elitist mutation-only evolutionary algorithms. Here
and below, by the runtime we mean the expected number of fitness
evaluations made until an optimum is found for the first time.
Upper bounds of the runtime of non-elitist GAs, involving the
crossover operators, were obtained later
in~\citep{bib:cdel14,er2016_proc}. The runtime bounds presented
in~\citep{bib:cdel14,bib:l11} are based on the drift analysis.
In~\citep{bib:ms15}, a runtime result is proposed for a class of
convex search algorithms, including some non-elitist
crossover-based GAs without mutation, on the so-called concave
fitness landscapes.

In this paper, we consider the non-elitist evolutionary algorithm
which uses a tournament selection and a mutation operator but no
crossover. The $s$-tournament selection randomly chooses $s$
individuals from the existing population and selects the best one
of them (see e.g. \citep{TG}). The mutation operator is viewed as
a randomized procedure, which computes one offspring with a
probability distribution depending on
the given parent individual. In this paper, evolutionary
algorithms with such outline are denoted as~EA. We study the
probability distribution of the EA population w.r.t. a set of
fitness levels.  The estimates of the EA behavior are based on
a~priori known parameters of a mutation operator. Using the
proposed model we obtain upper and lower bounds on expected
proportion of the individuals with fitness above certain
thresholds. The lower bounds are formulated in terms of linear
algebra and resemble the bound in Schemata
Theorem~\citep{bib:Gold89}. Instead of schemata here we consider
the sets of genotypes with the fitness bounded from below. Besides
that, the bounds obtained in this paper may be applied recursively
up to any given iteration.

A particular attention in this paper is payed to a special case
when mutation is {\em monotone}. Informally speaking, a mutation
operator is monotone if throughout the search space the following
condition holds: the greater the fitness of a parent the
``better'' offspring distribution the mutation generates.
One of the most well-known examples of monotone mutation is the
bitwise mutation in the case of \onemax fitness function. As shown
in~\citep{BE08}, in the case of monotone mutation, one of the most
simple evolutionary algorithms, known as the \oneplusoneEA has the
best-possible performance in terms of runtime and probability of
finding the optimum.

In the case of monotone mutation, the lower bounds on expected
proportions of the individuals turn into equalities for the
trivial evolutionary algorithm~\oneoneEA. This implies that the
tournament selection at least has no negative effect on the EA
performance in such a case. This observation is complemented by
the asymptotic analysis of the EA with monotone mutation
indicating that, given a sufficiently large population size and
some technical conditions, increasing the tournament size~$s$
always improves the EA performance.

As corollaries of the general lower bounds on expected proportions
of sufficiently fit individuals, we obtain polynomial upper bounds
on the Randomized Local Search runtime on unimodal functions and
upper bounds on runtime of EAs on \twosat problem and on a family
of Set Cover problems proposed by~\cite{Ba84}. Unlike the upper
bounds on runtime of evolutionary algorithms with tournament
selection from~\citep{bib:cdel14,er2016_proc,bib:l11}, which
require sufficiently large tournament size, the upper bounds on
runtime obtained here hold for any tournament size.

The rest of the paper is organized as follows. In
Section~\ref{sec:model}, we give a formal description of the
considered EA, introduce an approximating model of the EA
population and define some required parameters of the probability
distribution of a mutation operator in terms of fitness levels. In
Section~\ref{sec:bounds}, using the model from
Section~\ref{sec:model}, we obtain lower and upper bounds on
expected proportions of genotypes with fitness above some given
thresholds. Section~\ref{sec:monotone} is devoted to analysis of
an important special case of monotone mutation operator, where the
bounds obtained in the previous section become tight or
asymptotically tight. In Section~\ref{sec:app}, we consider some
illustrative examples of monotone mutation operators and
demonstrate some applications of the general results from
Section~\ref{sec:bounds}. In particular, in this section we obtain
new lower bounds for probability to generate optimal genotypes at
any given iteration~$t$ for a class of unimodal functions, for
\twosat problem and for a family of set cover problems proposed by
E.~Balas (in the latter two cases we also obtain upper bounds on
the runtime of the EA). Besides that in Section~\ref{sec:app} we
give an upper bound on expected proportion of optimal genotypes
for \onemax fitness function. Section~\ref{sec:conc} contains
concluding remarks.

This work extends the conference paper~\citep{Eremeev2000}. The
extension consists in comparison of the EA behavior to that of the
\oneoneEA, the \onelambdaEA and the \oneplusoneEA in
Section~\ref{sec:bounds} and in the new runtime bounds and tail
bounds demonstrated in Section~\ref{sec:app}. The main results
from the conference paper are refined and provided with more
detailed proofs.

\section{Description of Algorithms and Approximating Model} \label{sec:model}

\subsection{Notation and Algorithms} \label{subsec:algorithms}

Let the optimization problem consist in maximization of an
objective function~$f$ on the set of feasible solutions~${\rm Sol}
\subseteq {\mathcal X} =\{0,1\}^n$, where~${\mathcal X}$ is the
search space of all binary strings of length~$n$.

\paragraph{The Evolutionary Algorithm~EA.}
The EA searches for the optimal or sub-optimal solutions using a
population of individuals, where each individual (genotype)~$g$ is
a bitstring~$(g^1, g^2,\ldots,g^n)$, and its components~$g^i\in
\{0,1\}, i=1,2,...,n,$ are called genes.

In each iteration the EA constructs a new population on the basis
of the previous one. The search process is guided by the values of
a fitness function
$$  \phi(g) = \left\{
\begin{array}{ll}
f(g) &  \mbox{\rm if } \ \ g \in {\rm Sol};\\
r(g) &  \mbox{\rm otherwise,}
\end{array} \right.
$$
where $r(\cdot)$ is a penalty function.

The individuals of the population may be ordered according to the
sequence in which they are generated, thus the population may be
considered as a vector of genotypes
$X^t=(g_1^{(t)},g_2^{(t)},...,g_{\lambda}^{(t)})$, where
${\lambda}$ is the size of population, which is constant during
the run of the EA, and $t$ is the number of the current iteration.
In this paper, we consider a non-elitist algorithmic outline,
where all individuals of a new population are generated
independently from each other with identical probability
distribution depending on the existing population only.

Each individual is generated through selection of a parent
genotype by means of a selection operator, and modification of
this genotype in mutation operator. During the mutation, a subset
of genes in the genotype string $g$ is randomly altered. In
general the mutation operator may be viewed as a random variable
${\rm Mut}(g) \in {\mathcal X}$ with the probability distribution
depending on~$g$.

The genotypes of the initial population $X^0$ are generated with
some a priori chosen probability distribution. The stopping
criterion may be e.g. an upper bound on the number of
iterations~$t_{\max}$. The result is the best solution generated
during the run.
The EA has the following scheme. \vspace{0.5em}

\noindent 1. Generate the initial population $X^{0}$.\\
2. For $t:=0$ to $t_{\max}-1$ do\\
\mbox{\hspace{2em}} 2.1. For $k:=1$ to ${\lambda}$ do\\
\mbox{\hspace{5em}} Choose a parent genotype~$g$ from
$X^{t}$ by $s$-tournament selection. \\
\mbox{\hspace{5em}} Add $g_k^{(t+1)}={\rm Mut}(g)$ to the
population $X^{t+1}$. \vspace{0.5em}



In theoretical studies, the evolutionary algorithms are usually
treated without a stopping criterion (see
e.g.~\citep{NeumannWitt2010}). Unless otherwise stated, in the EA
we will also assume that~$t_{\max}=\infty.$

Note that in the special case of the EA with~$\lambda=1$ we can
assume that~$s=1$, since the tournament selection has no effect in
this case.


\paragraph{\onelambdaEA and
\oneplusoneEA.} In the following sections we will also need a
description of two simple evolutionary algorithms, known as the
\onelambdaEA and the \oneplusoneEA.

The genotype of the current individual on iteration~$\tau$ of the
\onelambdaEA will be denoted by $b^{(\tau)}$, and in the
\oneplusoneEA it will be denoted by $x^{(\tau)}$. The initial
genotypes $b^{(0)}$ and $x^{(0)}$ are generated with some a priori
chosen probability distribution. The only difference between the
\onelambdaEA and the \oneplusoneEA consists in the method of
construction of an individual for iteration~$\tau+1$ using the
current individual of iteration~$\tau$ as a parent. In both
algorithms the new individual is built with the help of a
{mutation operator}, which we will denote by~${\rm Mut}'$. In case
of the \onelambdaEA, the mutation operator is independently
applied~$\lambda$ times to the parent genotype~$b^{(\tau)}$ and
out of~$\lambda$ offspring a single genotype with the highest
fitness value is chosen as~$b^{(\tau+1)}$. (If there are several
offspring with the highest fitness, the new
individual~$b^{(\tau+1)}$ is chosen arbitrarily among them.) In
the \oneplusoneEA, the mutation operator is applied
to~$x^{(\tau)}$ once. If $x={\rm Mut}'(x^{(\tau)})$ is such that
$\phi(x)>\phi(x^{(\tau)}),$ then $x^{(\tau+1)}:=x$; otherwise
$x^{(\tau+1)}:=x^{(\tau)}$.

\subsection{The Proposed Model}

The EA may be considered as a Markov chain  in a number of ways.
For example, the states of the chain may correspond to different
vectors of ${\lambda}$ genotypes that constitute the population
$X^t$ (see~\citep{Gu}). In this case the number of states in the
Markov chain is~$2^{n{\lambda}}$. Another model representing the
GA as a Markov chain is proposed in \citep{NV}, where all
populations which differ only in the ordering of individuals are
considered to be equivalent. Each state of this Markov chain may
be represented by a vector of $2^n$ components, where the
proportion of each genotype in the population is indicated by the
corresponding coordinate and the total number of states
is~${2^n+\lambda-1 \choose \lambda}$. In the framework of this
model, M.Vose and collaborators have obtained a number of general
results concerning the emergent behavior of GAs by linking these
algorithms to the infinite-population GAs~\citep{Vo}.

The major difficulties in application of the above mentioned
models to the analysis of GAs for combinatorial optimization
problems are connected with the necessity to use the high-grained
information about fitness value of each genotype. In the present
paper, we consider one of the ways to avoid these difficulties by
means of grouping the genotypes into larger classes on the basis
of their fitness.



Assume that $\phi_0:=\min\{\phi(g): g \in {\mathcal X}\}$ and
there are $m$ level lines of the fitness function fixed such that
$\phi_0<\phi_1 < \phi_2 \ldots < \phi_m$. The number of levels and
the fitness values corresponding to them may be chosen
arbitrarily, but they should be relevant to the given problem and
the mutation operator to yield a meaningful model. Let us
introduce the sequence of Lebesgue subsets of~${\mathcal X}$
$$
H_i:=\{g : \phi(g) \geq \phi_i \}, \hspace{1em} i=0,\ldots,m.
$$
Obviously, $H_0={\mathcal X}$. For the sake of convenience, we
define $H_{m+1}:=\emptyset$. Also, we denote the level sets
$A_i:=H_i \backslash H_{i+1}, \ i=0,\ldots,m$ which give a
partition of~${\mathcal X}$. Partitioning subsets~$A_i$ are more
frequently used in literature on level-based analysis, compared to
the Lebesgue subsets~$H_i$. In this paper we will frequently state
that a genotype has a sufficiently high fitness, therefore the use
of subsets $H_i=\cup_{j=i}^m A_i$ will be more convenient in such
cases. One of the partitions used in the literature, called the
{\em canonical partition}, defines $\phi_0,\dots,\phi_m$ as the
set of all fitness values on the search space~$\mathcal X$.

\begin{figure}[htbr]
\begin{center}
\includegraphics[height=4.56cm, width=7.44cm]{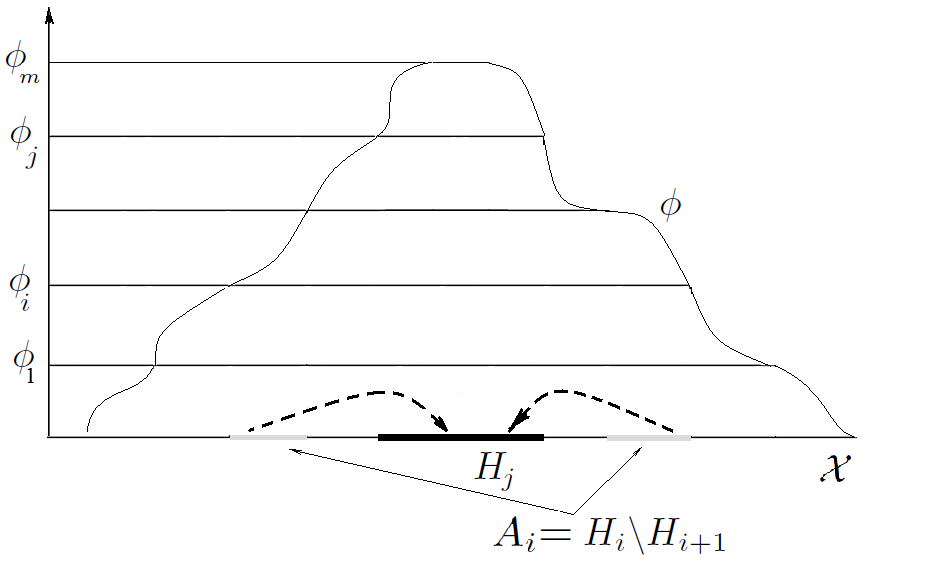}
\caption{Transitions from $A_i$ to $H_j$ under mutation.}
\label{fig:transitions}
\end{center}
\end{figure}

Now suppose that for all $i=0,...,m$ and $j=1,...,m,$ the a priori
lower bounds $\alpha_{ij}$ and upper bounds $\beta_{ij}$ on
mutation transition probabilities from subset $A_i$ to $H_j$ are
known, i.e.
$$
\alpha_{ij} \leq \Pr\{{\rm Mut}(g)\in H_j\} \leq \beta_{ij} \ \
\mbox{for any $g \in A_i$}.
$$
Fig.~\ref{fig:transitions} illustrates the transitions considered
in this expression.

Let $\A$ denote the matrix with the elements $\alpha_{ij}$ where
$i=0,...,m$, and $j=1,...,m$. The similar matrix of upper bounds
$\beta_{ij}$ is denoted by $\B$. Let the population on
iteration~$t$ be represented by the {\it population vector}
$$
\z^{(t)}=(z_1^{(t)},z_2^{(t)},\ldots,z_m^{(t)})
$$
where $z_i^{(t)}\in [0,1]$ is the proportion of genotypes from
$H_i$ in population~$X^t$. The population vector $\z^{(t)}$ is a
random vector, where $z_i^{(t)}\geq z_{i+1}^{(t)}$ for
$i=1,...,m-1$ since $H_{i+1} \subseteq H_i$.

Let $\Pr\{g^{(t)} \in H_j\}$ be the probability that an
individual, which is added after selection and mutation into
$X^t$, has a genotype from $H_j$ for $j=0,...,m$, and $t>0.$
According to the scheme of the EA this probability is identical
for all genotypes of $X^t$, i.e. $\Pr\{g^{(t)}\in
H_j\}=\Pr\{g_1^{(t)}\in H_j\}=...=\Pr\{g_{\lambda}^{(t)}\in
H_j\}$.
\begin{proposition} \label{vose}
$\E[z_i^{(t)}]=\Pr\{g^{(t)} \in H_i\}$ for all $t > 0, i=1,...,m$.
\end{proposition}
{\bf Proof.} Consider the sequence of identically distributed
random variables $\xi_1^i,\xi_2^i,...,\xi_{\lambda}^i$, where
$\xi_l^i=1$ if the $l$-th individual in the population $X^{t}$
belongs to~$H_i$, otherwise $\xi_l^i=0$. By the definition,
$z_i^{(t)}=\sum_{l=1}^{\lambda} \xi_l^i/{\lambda}$, consequently
$\E[z_i^{(t)}]=\sum_{l=1}^{\lambda} {\E[\xi_l^i]}/{\lambda}=
\sum_{l=1}^{\lambda} {\Pr\{g^{(t)} \in
H_i\}}/{\lambda}=\Pr\{g^{(t)} \in H_i\}. $ $\Box$

\paragraph{Level-Based Mutation.}
If for some mutation operator there exist two equal matrices of
lower and upper bounds~$\A$ and $\B$, i.e.  $\alpha_{ij} =
\beta_{ij}$ for all $i=0,\dots,m,$ $j=1,\dots,m$ then the mutation
operator will be called {\it level-based}. By this definition, in
the case of level-based mutation, ${\Pr\{{\rm Mut}(g)\in H_j\}}$
does not depend on a choice of genotype $g \in A_i$ and the
probabilities $\gamma_{ij}=\Pr\{{\rm Mut}(g)\in H_j|g \in A_i\}$
are well-defined. In what follows, we call~$\gamma_{ij}$ a {\it
cumulative transition probability}. The symbol~$\G$ will denote
the matrix of cumulative transition probabilities of a level-based
mutation operator.

If the EA uses a level-based mutation operator, then the
probability distribution of population~$X^{t+1}$ is completely
determined by the vector~$\z^{(t)}$. In this case the EA may be
viewed as a Markov chain with states corresponding to the elements
of
$$
Z_{\lambda}:=\left\{\z \in
\{0,{1}/{\lambda},{2}/{\lambda},\ldots,1\}^m : z_i\geq z_{i+1},
i=1\dots,m-1\right\},
$$
which is the set of all possible vectors of population of
size~${\lambda}$. Here and below, the symbol~$\z$ is used to
denote a vector from the set of all possible population
vectors~$Z_{\lambda}$.

The cardinality of set~$Z_{\lambda}$ may be evaluated analogously
to the number of states in the model of~\cite{NV}. Now levels
replace individual elements of the search space, which gives a
total of~${m+\lambda-1 \choose \lambda}$ possible population
vectors.

\section{Bounds on Expected Proportions of Fit Individuals}
\label{sec:bounds}

In this section, our aim is to obtain lower and upper bounds on
$\E[\z^{(t)}]$ for arbitrary~$s$ and $t$ if the distribution of
the initial population is known.

Let $P_{ch}(S,\z)$ denote the probability that the genotype,
chosen by the tournament selection from a population with
vector~$\z$, belongs to a subset $S \subseteq {\mathcal X}$. Note
that if the current population is represented by the vector
$\z^{(t)}=\z$, then a genotype obtained by selection and mutation
would belong to $H_j$ with a conditional probability
\begin{equation}
\label{t} \Pr\{g^{(t+1)} \in H_j |\z^{(t)}=\z\}=
\sum\limits_{i=0}^m \hspace{0.5em}\sum\limits_{g \in A_i}
\Pr\{{\rm Mut}(g) \in H_j |g \} P_{ch}(\{g\},\z).
\end{equation}

\subsection{Lower Bounds} \label{subsec:lower_bounds}

Expression (\ref{t}) and the definitions of bounds $\alpha_{ij}$
yield for all $j=1,\dots,m$:
\begin{equation}\label{eq:conditional}
\Pr\{g^{(t+1)} \in H_j |\z^{(t)}=\z\}\geq \sum\limits_{i=0}^m
\alpha_{ij} \sum\limits_{g \in A_i} P_{ch}(\{g\},\z) =
\sum\limits_{i=0}^m \alpha_{ij} P_{ch}(A_i,\z),
\end{equation}
which turns into an equality in the case of level-based mutation
and $\A=\G$.

Given a tournament size $s$ we obtain the following selection
probabilities: $P_{ch}(H_i,\z^{(t)})= 1-(1-z_i^{(t)})^s, \
i=1,\dots,m$, and, consequently, $P_{ch}(A_i,\z)=
{(1-z_{i+1})^s}-{(1-z_i)^s}$. This leads to the inequality:
$$
\Pr\{g^{(t+1)} \in H_j |\z^{(t)}=\z\} \geq\sum\limits_{i=0}^m
\alpha_{ij}((1-z_{i+1})^s-(1-z_i)^s).
$$
By the total probability formula,
\begin{equation}
\label{basic_geq_start} \Pr\{g^{(t+1)} \in H_j\}= \sum \limits_{\z
\in Z_{\lambda}}\Pr\{g^{(t+1)} \in
H_j|\z^{(t)}=\z\}\Pr\{\z^{(t)}=\z\}
\end{equation}
$$
\geq \sum \limits_{\z \in Z_{\lambda}}\sum\limits_{i=0}^m
\alpha_{ij} ((1-z_{i+1})^s-(1-z_i)^s)\Pr\{\z^{(t)}=\z\}
$$
$$
 =\sum\limits_{i=0}^m
\alpha_{ij}\E[(1-z_{i+1}^{(t)})^s-(1-z_{i}^{(t)})^s]
$$
\begin{equation}
\label{basic_geq} = \alpha_{mj} \E[(1-z_{m+1}^{(t)})^s]
 -\alpha_{0j} \E[(1-z_{0}^{(t)})^s] -
\sum\limits_{i=1}^m (\alpha_{ij}-\alpha_{i-1,j})
\E[(1-z_{i+1}^{(t)})^s],
\end{equation}
where the last expression is obtained by regrouping the summation
terms. Proposition~\ref{vose} implies that
$\E[z_j^{(t+1)}]=\Pr\{g^{(t+1)} \in H_j\}$. Consequently, since
$(1-z_{m+1}^{(t)})^s=1$ and $(1-z_{0}^{(t)})^s=0$,
expression~(\ref{basic_geq}) gives a lower bound
\begin{equation} \label{simple_expect}
\label{simple} \E[z_j^{(t+1)}] \geq
\alpha_{mj}-\sum\limits_{i=1}^m (\alpha_{ij}-\alpha_{i-1,j})
\E[(1-z_{i}^{(t)})^s].
\end{equation}

Note that~(\ref{simple_expect}) turns into an equality in the case
of level-based mutation and~$\A=\G$. We would like to
use~(\ref{simple_expect}) recursively $t$ times in order to
estimate $\E[{\bf z}^{(t)}]$ for any~$t$, given the initial vector
$\E[{\bf z}^{(0)}]$. It will be shown in the sequel that such a
recursion is possible under monotonicity assumptions defined
below.

\paragraph{Monotone Matrices and Mutation Operators.}
In what follows, any $((m+1)\times m)$-matrix~$\bf \Delta$ with
elements $\delta_{ij}, i=0,...,m$, $j=1,...,m,$ will be called
{\it monotone} iff $\delta_{i-1,j} \leq \delta_{ij}$ for all $i,j$
from 1 to $m$. Monotonicity of a matrix of bounds on transition
probabilities means that the greater fitness level~$A_i$ a parent
solution has, the greater is its bound on transition probability
to any subset $H_j,$ $j=1,\dots,d$. Note that for any mutation
operator the monotone upper and lower bounds exist. Formally, for
any mutation operator a valid monotone matrix of lower bounds
would be $\A={\bf 0}$ where ${\bf 0}$ is a zero matrix. A monotone
matrix of upper bounds, valid for any mutation operator is $\B=\bf
U$, where $\bf U$ is the matrix with all elements equal~1. These
are extreme and impractical examples. In reality a problem may be
connected with the absence of bounds which are sharp enough to
evaluate the mutation operator properly.

If given some set of levels $\phi_1, \ldots \phi_m,$ there exist
two matrices of lower and upper bounds~$\A,\B$ such that $\A=\B$
and these matrices are monotone then operator~${\rm Mut}$ is
called {\em monotone w.r.t. the set of levels} $\phi_1, \ldots,
\phi_m$. In this paper, we will also call such operators {\em
monotone} for short. Note that by the definition, any monotone
mutation operator is level-based, since $\alpha_{ij} = \beta_{ij}$
for all $i,$ $j$. The following proposition shows how the
monotonicity property may be equivalently defined in terms of
cumulative transition probabilities.

\begin{proposition} \label{prop:monotone}
A mutation operator~${\rm Mut}$ is monotone w.r.t. the set of
levels $\phi_1, \ldots \phi_m$ iff for any $i,i', j \in \{
0,\dots,m\},$ such that $i\ge i',$ for any genotypes $g\in A_i,$
$g'\in A_{i'}$ holds
$$
\Pr\{{\rm Mut}(g)\in H_j\}\geq \Pr\{{\rm Mut}(g')\in H_j \}.
$$
\end{proposition}

{\bf Proof.} Indeed, suppose that $\A=\B$ and these matrices are
monotone. Then for any genotypes $g\in A_i$ and $g'\in A_{i'}$,
$i\ge i'$ holds
$$
\Pr\{{\rm Mut}(g)\in H_j\} \geq \alpha_{ij} \ge
\alpha_{i'j}=\beta_{i'j} \geq \Pr\{{\rm Mut}(g') \in H_j\}.
$$

Conversely, if for any level~$j$ and any genotypes $g\in A_i$ and
$g'\in A_{i'}$, ${i\ge i'}$ holds $\Pr\{{\rm Mut}(g)\in H_j\}\geq
\Pr\{{\rm Mut}(g')\in H_j\}$, then taking $i=i'$ we note that
${\Pr\{{\rm Mut}(g)\in H_j\}}$ is equal for all $g\in A_i$ and one
can assign $\alpha_{ij}=\beta_{ij}={\Pr\{{\rm Mut}(g) \in H_j \ |
\ g\in A_i\}.}$ The resulting matrices $\A$ and $\B$ are obviously
monotone. $\Box$\\

Proposition~\ref{prop:monotone} implies that in the case of the
canonical partition, i.e. when $\{\phi_0, \phi_1, \ldots \phi_m\}$
is the set of all values of~$\phi(\cdot)$, operator~${\rm Mut}$ is
monotone w.r.t. $\phi_1, \ldots \phi_m$ iff for any genotypes $g$
and $g',$ such that $\phi(g)\geq \phi(g')$, for any $r\in \R$
holds
$$
\Pr\{\phi({\rm Mut}(g))\geq r\}\geq \Pr\{\phi({\rm Mut}(g'))\geq
r\}.
$$
The monotonicity of mutation operator w.r.t. a canonical partition
is equivalent to the definition of monotone reproduction operator
from~\citep{BE01} in the case of single-parent, single-offspring
reproduction. According to the terminology of~\cite{Daley}, such
random operators are also called {\em stochastically monotone}.

As a simple example of a monotone mutation operator we can
consider a {\em point mutation operator}: with probability~$q>0$
keep the given genotype unchanged; otherwise (with
probability~$1-q$) choose $i$ randomly from $\{1,\dots,n\}$ and
change gene~$i$. As a fitness function we take the function ${\rm
\onemax}(g) \equiv \sum_{i=1}^n |g_i|$, where $g\in\{0,1\}^n$. Let
us assume $m=n$ and define the thresholds $\phi_0:=0,
\phi_1=1,...,\phi_{n}=n$. All genotypes with the same fitness
function value have equal probability to produce an offspring with
any required fitness value, therefore this is a case of
level-based mutation. In such a case identical matrices of lower
and upper bounds $\A$ and $\B$ exist and they both equal to the
matrix of cumulative transition probabilities~$\G$. The latter
consists of the following elements: $\gamma_{ij}=1$ for all
$i={1,\dots,n}, \ j={0,\dots,i-1}$, since point mutation can not
reduce the fitness by more than one level; $\gamma_{i,i+1} =
(1-q)(n-i)/n$ for $i=0,\dots,n-1$ because with probability
$(1-q)(n-i)/n$ any genotype is upgraded;
$$
\gamma_{ii} = \cases{
 q+ \gamma_{i,i+1} & \mbox{\rm if }
$i=1,\dots,n-1$;
 \cr
q & \mbox{\rm if } $i=n$;
 }
$$
because a genotype in~$H_i$ can be obtained as an offspring of a
genotype from~$A_i$ in two ways: either the parent genotype has
been upgraded (which happens with probability~$\gamma_{i,i+1}$) or
it stays at level~$i$, which happens with probability~$q$; finally
$\gamma_{ij} = 0,$ $i=0,\dots,n-2, \ j=i+2,\dots,n$ because point
mutation can not increase the level number by more than~1. The
elements of matrix~$\G$ obviously satisfy the monotonicity
condition~$\gamma_{ij}-\gamma_{i-1,j}\ge 0$ when~$i\ne j$. For the
case of~$i=j$ we have~$\gamma_{ii}-\gamma_{i-1,i}=q+(q-1)/n$ which
is nonnegative if ${q}\ge 1/(n+1).$ Therefore with any ${q}\ge
1/(n+1),$ the matrix~$\G$ is monotone in this example and the
mutation operator is monotone as well.

\begin{proposition} \label{prop:c1}
If $\A$ is monotone, then for any tournament size $s\geq 1$ and
$j=1,\ldots,m$ holds
\begin{equation}
\label{c1} \E[z_j^{(t+1)}] \geq \alpha_{0j}+\sum\limits_{i=1}^m
(\alpha_{ij}-\alpha_{i-1,j})\E[z_{i}^{(t)}],
\end{equation}
besides that~(\ref{c1}) is an equality if $s=1$, operator~${\rm
Mut}$ is monotone and $\A$ is its matrix of cumulative transition
probabilities.
\end{proposition}

{\bf Proof.} Monotonicity of matrix~$\A$ implies that
$\alpha_{ij}-\alpha_{i-1,j}\ge 0$ for all $i=1,\dots,m,$
$j=1,\dots,m,$ so the simple estimate $(1-z_{i}^{(t)})^s \leq
1-z_{i}^{(t)}$ may be applied to all terms of the sum
in~(\ref{simple_expect}) and we get
$$
\E[z_j^{(t+1)}] \geq \alpha_{mj}-\sum\limits_{i=1}^m
(\alpha_{ij}-\alpha_{i-1,j}) (1-\E[z_{i}^{(t)}]).
$$
Regrouping the terms in the last bound we obtain the required
inequality~(\ref{c1}).

Finally, note that lower bound~(\ref{simple_expect}) holds as an
equality if the mutation operator is monotone and~$\A=\G$,
therefore the last lower bound is an equality in the case of
monotone~$\A=\G$ and~$s=1$. $\Box$


\paragraph{Lower Bounds from Linear Algebra.}

Let ${\bf W}$ be a $(m \times m)$-matrix with elements
$w_{ij}=\alpha_{ij} - \alpha_{i-1,j},$ let ${\bf I}$ be the
identity matrix of the same size, and denote $\alpha =
(\alpha_{01},...,\alpha_{0m})$. With these notations,
inequality~(\ref{c1}) takes a short form $\E[\z^{(t+1)}] \geq
\alpha+\E[\z^{(t)}] \ \W.$ Here and below, the inequality
sign~"$\le$" for some vectors $\x=(x_1,\dots,x_m)$ and
$\y=(y_1,\dots,y_m)$ means the component-wise comparison, i.e. $\x
\le \y$ iff $x_i \le y_i$ for all~$i$. The following theorem gives
a component-wise lower bound on vector~$\E[\z^{(t+1)}]$ for
any~$t$.

\begin{theorem} \label{th:straight}
Suppose that $||\cdot||$ is some matrix norm. If matrix $\A$ is
monotone and ${\lim\limits_{t \to \infty} ||\W^t||= 0}$, then for
all $t\geq 1$ holds
\begin{equation}
\label{c2} \E[\z^{(t)}]\geq \E[\z^{(0)}] {\bf W}^t+\alpha({\bf
I}-{\bf W})^{-1}({\bf I}-{\bf W}^t)
\end{equation}
and inequality~(\ref{c2}) turns into an equation if the tournament
size $s=1$, the mutation operator used in the EA is monotone and
$\A$ is its matrix of cumulative transition probabilities.
\end{theorem}

The proof of this theorem is similar to the well-known inductive
proof of the formula~${S_t=a(1-w)^{-1}(1-w^t),} \ w\in \R, \ a\in
\R,$ for a sum of terms $a_1,\dots,a_t$ in a geometric
series~${a_t=a w^{t-1}}.$ Note that the recursion $\E[\z^{(t+1)}]
\geq \alpha+\E[\z^{(t)}] \W$ is similar to the recursive formula
$S_{t+1}=a + S_t w,$ assuming~$S_0=0$. However in our case
matrices and vectors replace numbers, we have to deal with
inequalities rather than equalities and the initial
element~$\E[\z^{(0)}]$ may be non-zero unlike~$S_0$.\\

{\bf Proof of Theorem~\ref{th:straight}.} Let us consider a
sequence of $m$-dimensional vectors {\nolinebreak
$\uu^{(0)},\uu^{(1)},...,\uu^{(t)},...$,} where
$\uu^{(0)}=\E[\z^{(0)}]$, $\uu^{(t+1)}=\alpha+\uu^{(t)}{\bf W}$.
We will show that $\E[\z^{(t)}] \geq \uu^{(t)}$ for any $t$, using
induction on $t$. Indeed, for $t=0$ the inequality holds by the
definition of~$\uu^{(0)}$. Now note that the right-hand side
of~(\ref{c1}) will not increase if the components of
$\E[\z^{(t)}]$ are substituted with their lower bounds. Therefore,
assuming we already have ${\E[\z^{(\tau)}] \geq \uu^{(\tau)}}$ for
some~$\tau$ and substituting~$\uu^{(\tau)}$ for $\E[\z^{(\tau)}]$
we make an inductive step ${\E[\z^{(\tau+1)}] \geq
\uu^{(\tau+1)}}$.

By properties of the linear operators (see
e.g.~\citep{Kolmogorov}, Chapter~III,~\S~29), due to the
assumption that~$\lim\limits_{t \to \infty} ||\W^t||= 0,$ we
conclude that matrix $({\bf I}-{\bf W})^{-1}$ exists.

Now, using the induction on $t,$ for any $t\ge 1$ we will obtain
the identity
$$
\uu^{(t)}= \uu^{(0)} {\bf W}^t+\alpha({\bf I}-{\bf W})^{-1}({\bf
I}-{\bf W}^t)
$$
which leads to inequality~(\ref{c2}). Indeed, for the base case of
$\tau=1,$ by the definition of~$\uu^{(1)}$ we have the required
equality. For the inductive step, we use the following
relationship
$$
\uu^{(\tau+1)}= \uu^{(\tau)}{\bf W}+\alpha = \uu^{(0)} {\bf
W}^{\tau+1}+\alpha({\bf I}-{\bf W})^{-1} \left(\W - \W^{\tau+1}+
{\bf I}-\W \right) $$
$$
= \uu^{(0)} {\bf W}^{\tau+1}+\alpha({\bf I}-{\bf W})^{-1}({\bf
I}-{\bf W}^{\tau+1}). \ \ \Box
$$

In conditions of Theorem~\ref{th:straight}, the right-hand side
of~(\ref{c2}) approaches ${\alpha({\bf I}-{\bf W})^{-1}}$ when $t$
tends to infinity, thus the limit of this bound does not depend on
distribution of the initial population.

In many evolutionary algorithms, an arbitrary given genotype~$g'$
may be produced with a non-zero probability as a result of
mutation of any given genotype~$g$. Suppose that the probability
of such a mutation is lower bounded by some $\varepsilon>0$ for
all~$g,g'\in{\mathcal X}$. Then one can obviously choose some
monotone matrix~$\A$ of lower bounds that satisfies
$\alpha_{ij}\ge \varepsilon$ for all $i,j$. Thus,
$\alpha_{mj}-\alpha_{0j}\le 1-\varepsilon<1$ for all $j$. In this
case one can consider the matrix norm $||{\bf
W}||_{\infty}=\max_{j} \sum_{i=1}^m |w_{ij}|$. Due to monotonicity
of~$\A$ we have $w_{ij}=\alpha_{ij}-\alpha_{i-1,j}\geq 0$, so
$||{\bf W}||_{\infty}=\max_j \sum_{i=1}^m w_{ij}=\max_j
(\alpha_{mj}-\alpha_{0j})<1$, and the conditions of
Theorem~\ref{th:straight} are satisfied. A trivial example of a
matrix that satisfies the above description would be a matrix~$\A$
where all elements are equal to~$\varepsilon$.

Application of Theorem~\ref{th:straight} may be complicated due to
difficulties in finding the vector~$\alpha({\bf I}-{\bf W})^{-1}$
and in estimation the effect of multiplication by matrix~$\W^t.$
Some known results from linear algebra can help to solve these
tasks, as the example in Subsection~\ref{subsec:unimod} shows.
However sometimes it is possible to obtain a lower bound
for~$\E[\z^{(t)}]$ via analysis of the \oneoneEA algorithm,
choosing an appropriate mutation operator for it. This approach is
discussed below.

\paragraph{Lower Bounds from Associated Markov Chain.}

Suppose that a partition $A_0,\dots,A_m$ defined by
$\phi_0,\dots,\phi_m$ contains no empty subsets and let ${\T}$
denote a $(m+1) \times (m+1)$-matrix, with components
$$
t_{ij}=\alpha_{ij} - \alpha_{i,j+1}, \ i=0,\dots,m, \
j=0,\dots,m-1,
$$
$$
t_{im}=\alpha_{im}, \  i=0,\dots,m.
$$
Note that~$\T$ is a stochastic matrix so it may be viewed as a
transition matrix of a Markov chain, associated to the set of
lower bounds~$\alpha_{ij}$. This chain is a model of the
\oneoneEA, which is a special case of the \onelambdaEA
with~$\lambda=1$ (see Subsection~\ref{subsec:algorithms}). Suppose
that the \oneoneEA uses an artificial monotone mutation
operator~${\rm Mut}'$ where the cumulative transition
probabilities are defined by the bounds~$\alpha_{ij}$,
$i=0,\dots,m, \ j=1,\dots,m,$ corresponding to the EA mutation
operator~${\rm Mut}$. Namely, given a parent genotype~$x$, for any
$j=1,\dots,m$ we have $\Pr\{{\rm Mut}'(x)\in
A_j\}=\alpha_{ij}-\alpha_{i,j-1}$, where $i$ is such that $x\in
A_i$.
Operator~${\rm Mut}'(x)$ may be simulated e.g. by the following
two-stage procedure. At the first stage, a random index~$k$ of the
offspring level is chosen with the probability distribution
$\Pr\{k=j\}=\alpha_{ij}-\alpha_{i,j-1}, \ j=1,\dots,m,$ where $i$
is the level of parent~$x$. At the second stage, the offspring
genotype is drawn uniformly at random from~$A_k$. (Simulation of
the second stage may be computationally expensive for some fitness
functions but the complexity issues are not considered now.)
The initial search point~$b^{(0)}$ of the~\oneoneEA is generated
at random with probability distribution
defined by the probabilities $p^{(0)}_i:={\Pr\{\xi^{(0)} \in
A_i\}}= \E[z_i^{(0)}]-\E[z_{i+1}^{(0)}],$ $i=0,\dots,m$.
Denoting $\p^{(t)}:=\left(\Pr\{b^{(t)} \in
A_0\},\dots,\Pr\{b^{(t)} \in A_m\}\right)$,  by properties of
Markov chains we get $\p^{(t)}=\p^{(0)} \ {\bf T}^t$. The
following theorem is based on a comparison of~$\E[{\z}^{(t)}]$ to
the distribution of the Markov chain~$\p^{(t)}$.

\begin{theorem} \label{thm:T}
Suppose all level subsets $A_0,\dots,A_m$ are non-empty and
matrix~$\A$ is monotone. Then for any $t=1,2...$ holds
\begin{equation}\label{eq:fromMarkov}
\E[{\z}^{(t)}_i] \ge \p^{(0)} \ {\bf T}^t \ {\bf L},
\end{equation}
where ${\bf L}$ is a triangular $(m+1) \times (m+1)$-matrix with
components $\ell_{ij}=1$ if $i\ge j$ and $\ell_{ij}=0$ otherwise.
Besides that inequality~(\ref{eq:fromMarkov}) turns into an
equation if $s=1$, the EA mutation operator is monotone and $\A$
is its matrix of cumulative transition probabilities.
\end{theorem}

{\bf Proof.} The \oneoneEA described above is identical to an EA'
with~$\lambda=1$, $s=1$ and mutation operator~${\rm Mut}'$. Let us
denote the population vector of EA' by~$\hat{\z}^{(t)}$.
Obviously,
\begin{equation}\label{eq:x2z}
\hat{z}_i^{(t)}=\sum_{k=i}^m \Pr\{b^{(t)} \in A_k\}, \ \
i=1,\dots,m.
\end{equation}
Proposition~\ref{prop:c1} implies that in the original EA with
population size~$\lambda$ and tournament size~$s$, the
expectation~${\bf E}[\z^{(t)}]$ is lower bounded by the
expectation ${\bf E}[\hat{\z}^{(t)}]$ since~(\ref{c1}) holds as an
equality for the whole sequence of~${\bf E}[\hat{\z}^{(t)}]$ and
the right-hand side of~(\ref{c1}) is non-decreasing
on~$\E[z_{i}^{(t)}]$. Equality $\p^{(t)}=\p^{(0)} \ {\bf T}^t$
together with~(\ref{eq:x2z}) imply the required
bound~(\ref{eq:fromMarkov}). $\Box$

Note that inequalities~(\ref{c2}) and (\ref{eq:fromMarkov}) in
Theorems~\ref{th:straight} and~\ref{thm:T} turn into equalities if
these theorems are applied to the EA with $\lambda=1$ and monotone
mutation operator~${\rm Mut'}$ defined above. Therefore both
theorems guarantee equal lower bounds on~$\E[\z{(t)}]$, given
equal matrices~$\A$.

Subsections~\ref{subsec:2SAT} and~\ref{subsec:Balas} provide two
examples illustrating how Theorem~\ref{thm:T} may be used to
import known results on Markov chains behavior. The example from
Subsection~\ref{subsec:Balas} employs Theorem~\ref{thm:T} for
finding a vector~$\alpha({\bf I}-{\bf W})^{-1},$ so that
Theorem~\ref{th:straight} may be applied to bound~$\E[z_m^{(t)}]$
from below.

\subsection{Upper Bounds}

In this subsection, we obtain upper bounds on $\E[z_j^{(t+1)}]$
using a reasoning similar to the proof of
Proposition~\ref{prop:c1}. Expression~(\ref{t}) for all
$j=1,\dots,m$ yields:
\begin{equation}\label{eq:conditional1}
\Pr\{g^{(t+1)} \in H_j |\z^{(t)}=\z\}\leq \sum\limits_{i=0}^m
\beta_{ij} P_{ch}(A_i,\z)=\sum\limits_{i=0}^m
\beta_{ij}((1-z_{i+1})^s-(1-z_i)^s),
\end{equation}
which turns into equality in the case of level-based mutation. By
the total probability formula we have:

\begin{equation}
\label{basic_geq1} \E[z_j^{(t+1)}]= \sum \limits_{\z \in
Z_{\lambda}}\Pr\{g^{(t+1)} \in H_j|\z^{(t)}=\z\}\Pr\{\z^{(t)}=\z\}
\end{equation}
$$
\le \sum\limits_{i=0}^m
\beta_{ij}\E[(1-z_{i+1}^{(t)})^s-(1-z_{i}^{(t)})^s],
$$
 so
\begin{equation} \label{simpleabove}
\label{simple_above}
 \E[z_j^{(t+1)}]\leq \beta_{mj}-\sum\limits_{i=1}^m
(\beta_{ij}-\beta_{i-1,j})\E[(1-z_{i}^{(t)})^s].
\end{equation}

Under the expectation in the right-hand side we have a convex
function on~$z_{i}^{(t)}$. Therefore, in the case of monotone
matrix~$\B$, using Jensen's inequality  (see e.g.~\citep{Rudin},
Chapter~3) we obtain the following proposition.

\begin{proposition} \label{prop:upper_bound}
If $\B$ is monotone then
\begin{equation}
\label{t_above} \E[z_j^{(t+1)}] \leq \beta_{mj}-
\sum\limits_{i=1}^m
(\beta_{ij}-\beta_{i-1,j})(1-\E[z_{i}^{(t)}])^s.
\end{equation}
\end{proposition}


By means of iterative application of inequality~(\ref{t_above})
the components of the expected population vectors~$\E[\z^{(t)}]$
may be bounded up to arbitrary~$t$, starting from the initial
vector~$\E[\z^{(0)}]$. The nonlinearity in the right-hand side
of~(\ref{t_above}), however, creates an obstacle for obtaining an
analytical result similar to the bounds of
Theorems~\ref{th:straight} and \ref{thm:T}.

Note that all of the estimates obtained up to this point are
independent of the population size and valid for
arbitrary~${\lambda}$. In the Section~\ref{sec:monotone} we will
see that the right-hand side of~(\ref{t_above}) reflects the
asymptotic behavior of population under monotone mutation operator
as ${\lambda} \to \infty$.

\subsection{Comparison of EA to \onelambdaEA and \oneplusoneEA}
\label{subsec:compare}

This subsection shows how the probability of generating the
optimal genotypes at a given iteration of the EA relates to
analogous probabilities of \onelambdaEA and \oneplusoneEA.
The analysis here will be based on upper bound~(\ref{t_above}) and
on some previously known results provided in the attachment.

Suppose, matrix~$\B$ gives the upper bounds for cumulative
transition probabilities of the mutation operator~${\rm Mut}$ used
in the EA. Consider the \onelambdaEA and the \oneplusoneEA, based
on a monotone mutation operator~${\rm Mut}'$ for which~$\B$ is the
matrix of cumulative transition probabilities and suppose that the
initial solutions~$b^{(0)}$ and $x^{(0)}$ have the same
distribution over the fitness levels as the best incumbent
solution in the EA population~$X^0$. Formally: $\Pr\{{\rm
Mut}'(x)\in H_j\} = \beta_{ij}$ for any $x \in A_i,$ $i=0,\dots,m,
\ j=1,\dots,m,$ and $\Pr\{b^{(0)}\in H_j\}=\Pr\{x^{(0)}\in
H_j\}={\Pr\{\max_{k=1,\dots,\lambda} \phi(g^{(0)}_k) \ge
\phi_j\},} \ j=1,\dots,m.$ In what follows,  for any $j=1,\dots,m$
by $P^{(\tau)}_j$ we denote the probability that current
individual~$b^{(\tau)}$ on iteration~$\tau$ of the \onelambdaEA
belongs to~$H_j$. Analogously $Q^{(\tau)}_j$ denotes the
probability $\Pr\{x^{(\tau)}\in H_j\}$ for the \oneplusoneEA.

The following proposition is based on upper bound~(\ref{t_above})
and the results from~\citep{bor01,BE01} that allow to compare the
performance of the EA, the \onelambdaEA and the \oneplusoneEA.

\begin{proposition} \label{prop:upper_bound1}
Suppose that matrix~$\B$ is monotone. Then for any~$t\ge 0$ holds
$$
\E[z_m^{(t+1)}]\le \beta_{mm} - (\beta_{mm}-\beta_{m-1,m})
(1-P^{(t)}_m)^s \leq \beta_{mm}-
(\beta_{mm}-\beta_{m-1,m})(1-Q^{(t\lambda)}_m)^s.
$$
\end{proposition}

{\bf Proof.} Let us compare the EA to the \onelambdaEA and to the
\oneplusoneEA using the mutation and initialization procedures as
described above. Theorem~\ref{th:bor} (see the appendix) together
with Proposition~\ref{vose} imply that
$\E[z_{m}^{(t)}]=\Pr\{g^{(t)}\in H_m\} \le P^{(t)}_m$ for
all~$t\ge 0$. Furthermore, Theorem~5 from~\citep{BE01} (see the
appendix) implies that $P^{(t)}_m \le Q^{(t\lambda)}_m$ for
all~$t\ge 0$. Using Proposition~\ref{prop:upper_bound} and
monotonicity of~$\B$, we conclude that both claimed inequalities
hold.
$\Box$\\


\section{EA with Monotone Mutation Operator}
\label{sec:monotone}

First of all note that in the case of monotone mutation operator,
two equal monotone matrices of lower and upper bounds $\A=\B$
exist, so the bounds~(\ref{simple_expect}) and (\ref{simpleabove})
give equal results, and assuming $\G=\A=\B$ we get
\begin{equation} \label{eqn:equal_bounds}
\E[z_j^{(t+1)}] = \gamma_{mj}-\sum\limits_{i=1}^m
(\gamma_{ij}-\gamma_{i-1,j})\E[(1-z_{i}^{(t)})^s], \ \
j=1,\dots,m, \ \ t=0,1,\dots.
\end{equation}
This equality will be used several times in what follows.

In general, the population vectors are random values whose
distributions depend on~${\lambda}$. To express this in the
notation let us denote the proportion of genotypes from $H_i$ in
population~$X^t$ by $z^{(t)}_i({\lambda}), \ i=1,\dots,m$.

The following Lemma~\ref{third} and Theorem~\ref{th:asymptotics}
based on this lemma indicate that in the case of monotone
mutation, recursive application of the formula from right-hand
side of upper bound~(\ref{t_above}) allows to compute the expected
population vector of the infinite-population EA at any
iteration~$t$.

\begin{lemma} \label{third}
Let the EA use a monotone mutation operator with cumulative
transition probabilities matrix~$\G$, and let the genotypes of the
initial population be identically distributed. Then

(i) for all $t=0,1,... $ and $i=1,\dots,m$ holds
\begin{equation}
\label{ind} \lim \limits_{{\lambda}\to \infty} \left(
\E\left[\left(1-z_{i}^{(t)}({\lambda})\right)^s\right] -
 \left(1-\E[z_{i}^{(t)}({\lambda})]\right)^s \right)=0;
\end{equation}

(ii) if the sequence of $m$-dimensional vectors \linebreak
{\nolinebreak ${\bf u}^{(0)},{\bf u}^{(1)},...,{\bf u}^{(t)},...$}
is defined as
\begin{equation}\label{eqn:y_start}
{\bf u}^{(0)}=\E[{\z}^{(0)}({\lambda})],
\end{equation}
\begin{equation}\label{eqn:y_iter}
u^{(t+1)}_j=\gamma_{mj}- \sum\limits_{i=1}^m
(\gamma_{ij}-\gamma_{i-1,j})(1-u_{i}^{(t)})^s
\end{equation}
for $j=1,...,m$ and $t \geq 0$. Then $\lim \limits_{{\lambda} \to
\infty} \E[{\z}^{(t)}({\lambda})]= {\bf u}^{(t)}$ for all
$j=1,...,m$ at any iteration $t$.
\end{lemma}

The main step in the proof of Lemma~\ref{third}~(i) will consist
in showing that for a supplementary random variable
$X=(1-z_i^{(t)}({\lambda}))^s-(1-\E[z_i^{(t)}({\lambda})])^s,$ the
value of~$|\E[X]|$ is upper-bounded by an arbitrary
small~$\varepsilon>0$. This step is made by splitting the
range~$[-1,1]$ of~$X$ into a ``high-probability'' area and a
``low-probability'' area in such a way that $|X|$ is at
most~$\varepsilon$ in the ``high-probability'' area. Analogous
technique is used e.g. in the proof of Lebesgue Theorem, see
e.g.~\cite{Kolmogorov}, Chapter~VII,~\S~44.

{\bf Proof of Lemma~\ref{third}.} From~(\ref{eqn:equal_bounds}),
we conclude that if statement~(i) holds, then with ${\lambda} \to
\infty,$ the convergence of $\E[{\z}^{(t)}({\lambda})]$ to ${\bf
u}^{(t)}$ will imply that $\E[{\z}^{(t+1)}({\lambda})] \to {\bf
u}^{(t+1)}$. Thus, statement~(ii) follows by induction on~$t$.

Let us now prove statement~(i). Given some~$t,$ to
prove~(\ref{ind}) we recall the sequence of i.i.d. random
variables ${\mathcal I}_1^i,{\mathcal I}_2^i,...,{\mathcal
I}_{\lambda}^i$, where ${\mathcal I}_k^i=1$, if the $k$-th
individual of population $X^{t}$ belongs to~$H_i$, otherwise
${\mathcal I}_k^i=0$. By the law of large numbers, for any
$i=1,...,m$ and $\varepsilon
>0$, we have
$$
\lim \limits_{{\lambda} \to \infty}
 \Pr\left\{\left|
\frac{\sum_{k=1}^{\lambda} {\mathcal I}_k^i}{\lambda} -
\E[{\mathcal I}_1^i]\right|<\varepsilon\right\}
 = 1.
$$
Note that $\sum_{k=1}^{\lambda}{\mathcal
I}_k^i/{\lambda}=z_i^{(t)}({\lambda})$. Besides that, due to
Proposition~\ref{vose}, $\E[{\mathcal I}_1^i]={\Pr\{{\mathcal
I}_1^i=1\}}= \E[z_i^{(t)}({\lambda})].$ (In the case of $t=0$ this
equality holds as well, since all individuals of the initial
population are distributed identically.) Therefore, for any
$\varepsilon
>0$ the convergence $\Pr\left\{\left|z_i^{(t)}({\lambda})-\E[z_i^{(t)}({\lambda})]\right|<\varepsilon\right\}
\mathop{\longrightarrow} 1$ holds. Now by continuity of the
function $(1-x)^s$, it follows that
$$
\lim\limits_{{\lambda} \to \infty}
\Pr\left\{\left|(1-z_i^{(t)}({\lambda}))^s-(1-\E[z_i^{(t)}({\lambda})])^s\right|
\geq \varepsilon\right\} = 0.
$$
Let us denote ${\rm
F}_{\lambda}(x):=\Pr\left\{(1-z_i^{(t)}({\lambda}))^s-(1-\E[z_i^{(t)}({\lambda})])^s<x
\right\}$. Then
$$
\lim\limits_{{\lambda}\to \infty}
\left(\E\left[(1-z_i^{(t)}({\lambda}))^s\right]-(1-\E[z_i^{(t)}({\lambda})])^s
\right)= \lim\limits_{{\lambda}\to \infty}
\int\limits_{-\infty}^\infty x \, d{\rm F}_{\lambda}(x) \leq
$$
$$
\leq \lim\limits_{{\lambda}\to \infty}
\Pr\left\{\left|(1-z_i^{(t)}({\lambda}))^s-(1-\E[z_i^{(t)}({\lambda})])^s\right|
\geq \varepsilon\right\}+ \lim\limits_{{\lambda}\to
\infty}\int\limits_{|x|<\varepsilon} \varepsilon \, d{\rm
F}_{\lambda}(x)  \le \varepsilon
$$
for arbitrary $\varepsilon >0$, hence (\ref{ind}) holds. $\Box$\\

Combining equality~(\ref{eqn:equal_bounds}) with claim~(i) of
Lemma~\ref{third} we obtain a recursive expression
for~$\E[\z^{(t)}]$ in the infinite-population EA, which is
formulated as

\begin{theorem}\label{th:asymptotics}
If the mutation operator is monotone and individuals of the
initial population are distributed identically, then
\begin{equation} \label{eqn:assympt_bounds}
\lim\limits_{{\lambda} \to \infty} \E[z_j^{(t+1)}({\lambda})] =
\gamma_{mj}-\sum\limits_{i=1}^m
(\gamma_{ij}-\gamma_{i-1,j})(1-\E[z_{i}^{(t)}({\lambda})])^s
\end{equation}
for all $j=1,\dots,m,\ t \ge 0$.
\end{theorem}

For any $i,j$ and $t>0,$ the term $u_j^{(t)}$ of the sequence
defined by~(\ref{eqn:y_iter}) is nondecreasing in $u_i^{(t-1)}$
and in $s$ as well. With this in mind, we can expect that the
components of population vector of the infinite-population EA will
typically increase with the tournament size.
Theorem~\ref{th:tourn_size} below gives a rigorous proof of this
fact under some technical conditions on distributions of~$\rm Mut$
and~$X^0$.

\begin{theorem} \label{th:tourn_size}
Let ${\z}^{(t)}$ and $\hat{{\z}}^{(t)}$ correspond to EAs with
tournament sizes~$s$ and~$\hat{s}$, where ${s < \hat{s}}$. Besides
that, suppose that ${\rm Mut}$ is monotone with
$\gamma_{mj}>\gamma_{0j}$ for all~$j=1,\dots,m$ and the
individuals of initial populations are identically distributed so
that $\Pr\{g^{(0)}\in H_i\}\in (0,1)$ for all $i=1,\dots,m$. Then
for any $t>0$, given a sufficiently large~${\lambda},$ holds
$$
\E[\hat{z_i}^{(t)}(\lambda)] > \E[z_i^{(t)}(\lambda)], \ \ \
i=1,...,m.
$$
\end{theorem}

{\bf Proof.} Let the sequences $\{{\bf u}^{(t)}\}$ and $\{\hat{\bf
u}^{(t)}\}$ be defined as in Lemma~\ref{third}, corresponding to
tournament sizes~$s$ and $\hat{s}$. By the above assumptions,
${\bf u}^{(0)}=\hat{\bf u}^{(0)}.$

Now since $\Pr\{g^{(0)}\in H_i\}\in (0,1)$ for all $i=1,\dots,m,$
we have $u_{i}^{(0)}=\hat{u}_{i}^{(0)}\in(0,1)$ for
any~$i=1,\dots,m$. Thus, for all $j=1,\dots,m$ holds
\begin{equation}\label{eqn:first_step}
u^{(1)}_j=\gamma_{mj}- \sum\limits_{i=1}^m
(\gamma_{ij}-\gamma_{i-1,j})(1-u_{i}^{(0)})^s
 <
\gamma_{mj}- \sum\limits_{i=1}^m
(\gamma_{ij}-\gamma_{i-1,j})(1-\hat{u}_{i}^{(0)})^{\hat{s}}=\hat{u}^{(1)}_j,
\end{equation}
since $s<\hat{s}$ and $\gamma_{ij}-\gamma_{i-1,j}>0$ at least for
one of the levels~$i$ according to the assumption that
$\gamma_{mj}>\gamma_{0j}$. Due to the same reason, for all
$j=1,\dots,m$ from the last equality in~(\ref{eqn:first_step}) we
get ${\hat{u}^{(1)}_j<\gamma_{mj}\le 1.}$ Using the fact that
$(1-u_{i}^{(0)})^s \leq 1-u_{i}^{(0)}$ and re-arranging the terms
as in the proof of Proposition~\ref{prop:c1} we get
$$
 u^{(1)}_j
\geq \gamma_{0j}+\sum\limits_{i=1}^m (\gamma_{ij}-\gamma_{i-1,j})
\ u_{i}^{(0)}>0.
$$
To sum up, for $t=1$ we have $u_{i}^{(1)}<\hat{u}_{i}^{(1)},$
$u_{i}^{(1)}\in(0,1)$ and $\hat{u}_{i}^{(1)}\in(0,1)$.

Furthermore, if we assume that for all $i=1,\dots,m$ holds
$u_{i}^{(t-1)}<\hat{u}_{i}^{(t-1)},$ ${u_{i}^{(t-1)}\in(0,1)}$ and
$\hat{u}_{i}^{(t-1)} \in (0,1)$ then analogously
to~(\ref{eqn:first_step}) we get $u^{(t)}_j<\hat{u}^{(t)}_j$ for
all $j=1,\dots,m$. Besides that, just as in the case of~$t=1$ we
get $\hat{u}^{(t)}_j\in (0,1)$ and $u^{(t)}_j\in (0,1).$ So by
induction we conclude that $u^{(t)}_j<\hat{u}^{(t)}_j$ for all
$j=1,\dots,m$ and all $t>0$.

Finally, by claim~(ii) of Lemma~\ref{third}, for any~$i$ and $t$,
given a sufficiently large~$\lambda$, holds
$\E[\hat{z_i}^{(t)}({\lambda})]
> \E[z_i^{(t)}({\lambda})]$.  $\Box$\\

Informally speaking, Theorem~\ref{th:tourn_size} implies that in
the case of monotone mutation operator an optimal selection
mechanism consists in setting~$s\to \infty,$ which actually
converts the EA into the \onelambdaEA.

\section{Applications and Illustrative Examples}
\label{sec:app}


\subsection{Examples of Monotone Mutation Operators} \label{subsec:monot_ex}

Let us consider two cases where the mutation is monotone and the
matrices~$\G$ have a similar form.

First we consider the simple fitness function ${\rm \onemax}(g)$.
Suppose that the EA uses the bitwise mutation operator, changing
every gene with a given probability~$p_{\rm m}$, independently of
the other genes. Let the subsets $H_0,...,H_m$ be defined by the
level lines $\phi_0=0, \phi_1=1,...,\phi_m=m$ and $m=n$. The
matrix $\G$ for this operator could be obtained using the result
from~\citep{Ba}, but here we shall consider this example as a
special case of a more general setting.

Let the representation of the problem admit a decomposition of the
genotype string into $d$ non-overlapping substrings (called {\it
blocks} here) in such a way that the fitness function equals the
number of blocks for which a certain property~$\K$ holds. The
functions of this type belong to the class of additively
decomposed functions, where the elementary functions are Boolean
and substrings are non-overlapping (see e.g.~\citep{MMR}). Let
$K(g,\ell)=1$ if $\K$ holds for the block $\ell$ of genotype $g$,
and $K(g,\ell)=0$ otherwise (here $\ell=1,...,d$).

Suppose that during mutation, any block for which $\K$ did not
hold, gets the property~$\K$ with probability~$\tilde{r}$, i.e.
$$
{\Pr\{K({\rm Mut}(g),\ell)=1|K(g,\ell)=0\}=\tilde{r}}, \ \
\ell=1,...,d.
$$
On the other hand, assume that a block with the property $\K$
keeps this property during mutation with probability $r$, i.e.
$$
\Pr\{K({\rm Mut}(g),\ell)=1|K(g,\ell)=1\}=r, \ \ \ell=1,...,m.
$$
Let $m=d$ and the subsets $H_0,...,H_m$ correspond to the level
lines ${\phi_0=0,} {\phi_1=1,}...,{\phi_m=m}$ again. In this case
the element $\gamma_{ij}$ of cumulative transition probabilities
matrix~$\G$ equals the probability to obtain a genotype
containing~$j$ or more blocks with property~$\K$ after mutation of
a genotype which contained~$i$ blocks with this property. Let
$P(k',k)$ denote the probability that during mutation $k'$ blocks
without property $\K$ would produce~$k$ blocks with this property
and let $Q(i,l)$ denote the probability that after mutation of a
set of $i$ blocks with property $\K$, there will be at least~$l$
blocks with property $\K$ among them. (If $l>i$ then $Q(i,l):=0.$)
With these notations,
$$
\gamma_{ij}= \sum_{k=0}^{m-i} P(m-i,k) Q(i,j-k).
$$
Clearly, $P(k',k) = {k' \choose k}
\tilde{r}^k(1-\tilde{r})^{k'-k}$ and $
Q(i,l)=\sum_{\nu=0}^{\min\{i,i-l\}}{i \choose \nu} (1-r)^\nu
r^{i-\nu}. $ Thus,
\begin{equation}
\label{routine} \gamma_{ij}= \sum_{k=0}^{m-i} {m-i \choose k}
\tilde{r}^k (1-\tilde{r})^{m-i-k}
\sum_{\nu=0}^{\min\{i,i-(j-k)\}}{i \choose \nu} (1-r)^\nu
r^{i-\nu}.
\end{equation}
It is shown in~\citep{Eremeev2000,BE08} that if $r\geq \tilde{r}$
then matrix~$\G$ defined by~(\ref{routine}) is monotone.

Now matrix $\G$ for the bitwise mutation on \onemax function is
obtained assuming that $\tilde{r}=(1-r)=p_{\rm m}$ and $m=d=n$.
This operator is monotone in view of the above mentioned result,
if $p_{\rm m}\leq 0.5$, since in this case $r\geq \tilde{r}$. The
monotonicity of bitwise mutation on \onemax is used in works of
\cite{DJW10} and \cite{Witt_Linear}.

Expression~(\ref{routine}) may be also used for finding the
cumulative transition matrices of some other optimization problems
with a regular structure. As an example, below we consider the
vertex cover problem~(\vcp) on graphs of a special structure.

In general, the vertex cover problem is formulated as follows. Let
$G=(V,E)$ be a graph with a set of vertices
$V=\{v_1,\dots,v_{|V|}\}$ and the edge set
$E=\{e_1,\dots,e_{|E|}\}$ where $e_i=\{u(i),v(i)\}\subseteq V, \
i=1,\dots,|E|$. A subset $C \subseteq V$ is called a vertex cover
of $G$ if every edge has at least one endpoint in $C$. The vertex
cover problem is to find a vertex cover $C^{*}$ of minimal
cardinality.

Suppose that the \vcp is handled by the EA with the following
representation: each gene $g^i\in\{0,1\}, i=1,...,|E|$ corresponds
to an edge~$e_i$ of $G$, assigning one of its endpoints which has
to be included in the cover~$C(g)$. To be specific, we can assume
that~$g^i=1$ means that~$u(i)\in C(g)$ and $g^i=0$ means
that~$v(i)\in C(g)$. The vertices, not assigned by one of the
chosen endpoints, do not belong to~$C(g)$. On one hand, this
edge-based representation is degenerate in the sense that one
vertex cover~$C$ may be encoded by different genotypes~$g$. On the
other hand, any genotype~$g$ defines a feasible cover~$C(g)$. A
natural way to choose the fitness function in the case of this
representation is to assume $\phi(g)=|V|-|C(g)|$.

Note that most publications on evolutionary algorithms for \vcp
use the vertex-based representation with~$|V|$ genes, where
$g_j=1,\ j=1,\dots,|V|$ implies inclusion of vertex~$v_j$ into~$C$
(see e.g. \citep{NeumannWitt2010},~\S~12.1). In contrast to the
edge-based representation, the vertex-based representation is not
degenerate but some genotypes in this representation may define
infeasible solutions.

Following~\citep{Saiko} we denote by~$G(m)$ the graph consisting
of $m$ disconnected triangle subgraphs. Each triangle is covered
optimally by two vertices and the redundant cover consists of
three vertices. In spite of simplicity of this problem, it is
proven in~\citep{Saiko} that some well-known algorithms of branch
and bound type require exponential in~$m$ number of iterations if
applied to the \vcp on graph~$G(m)$.


In the case of $G(m)$, the fitness~$\phi(g)$ coincides with the
number of optimally covered triangles in~$C(g)$ (i.e. triangles
where only two different vertices are chosen), since covering
non-optimally all triangles gives~$C(g)=V$ and each optimally
covered triangle decreases the size of the cover by one. Let the
genes representing the same triangle constitute a single block,
and let the property~$\K$ imply that a triangle is optimally
covered. Then by looking at the two possible ways to produce a
gene triplet that redundantly covers a triangle, (i)~given a
redundant triangle and (ii)~given an optimally covered triangle,
we conclude that (i)~$\tilde{r}=1-p_{\rm m}^3-(1-p_{\rm m})^3$ and
(ii)~$r=1-p_{\rm m}(1-p_{\rm m})^2-p_{\rm m}^2(1-p_{\rm m})$.
Using~(\ref{routine}) we obtain the cumulative transition matrix
for this mutation operator. It is easy to verify that in this case
the inequality $r\geq \tilde{r}$ holds for any mutation
probability $p_{\rm m}$, and therefore the operator is always
monotone.

\begin{figure}[htbr]
\begin{center}
\includegraphics[height=5.71cm, width=11.25cm]{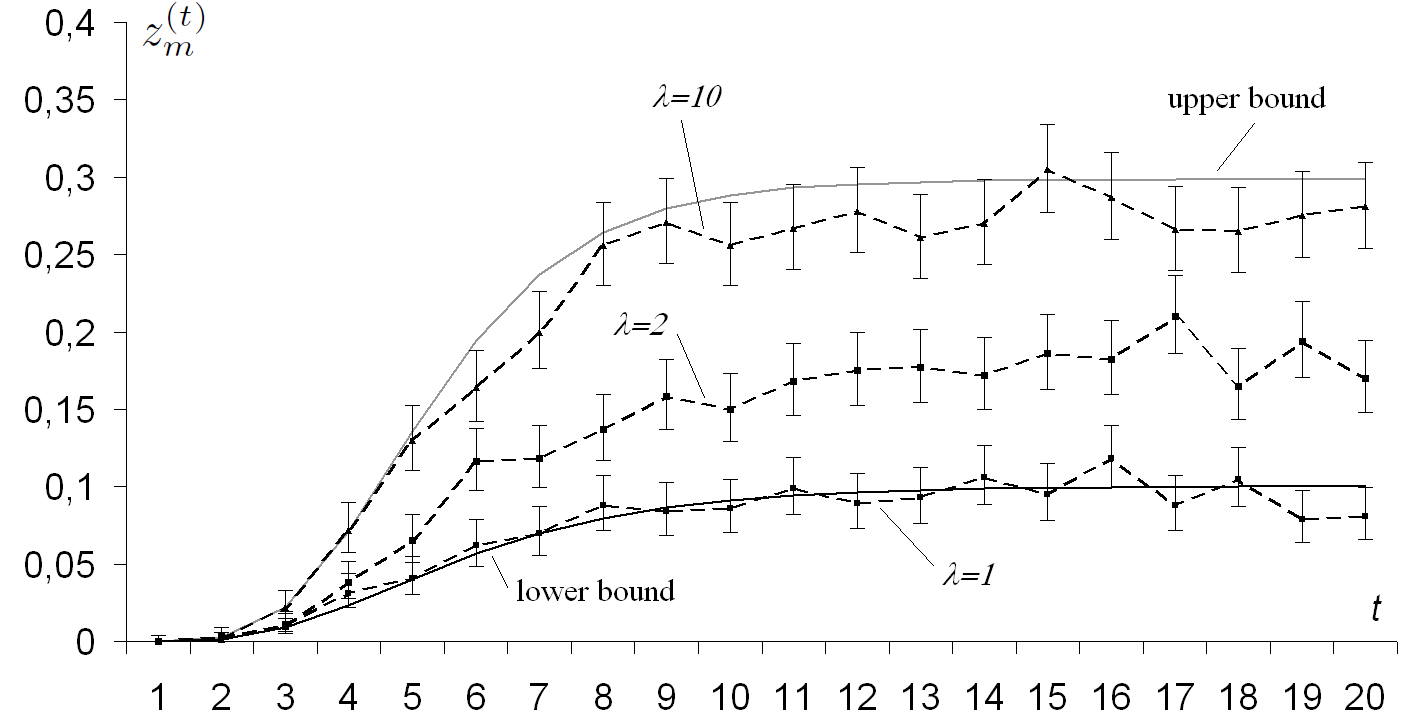}
\caption{Average proportion of optimal \vcp solutions and the
theoretical lower and upper bounds as functions of the iteration
number. Here $s=2$, $\lambda=1,2$ and~10.}
\label{fig:different_lambda}
\end{center}
\end{figure}

\paragraph{Computational Experiments.} Below we present some
experimental results in comparison with the theoretical estimates
obtained in Section~\ref{sec:bounds}. To this end we consider an
application of the EA to the \vcp on graphs~$G(m)$. The average
proportion of optimal genotypes in the population for different
population sizes is presented in
Figure~\ref{fig:different_lambda}. Here $m=8$, $p_{\rm m}=0.1$,
$s=2$ and $\z^{(0)}={\bf 0}$ (these parameters are chosen to
ensure clear visibility on plots). The statistics is accumulated
in 1000 independent runs of the algorithm where for each~$t$ only
one individual~$g_1^{(t)}$ was checked for optimality. Thus for
each~$t$ we have a series of 1000 Bernoully trials with a success
probability $\Pr\{g_1^{(t)}\in H_m\}=\E[z_m^{(t)}]$ which is
estimated from the experimental data. The 95\%-confidence
intervals for success probability in Bernoully trials are computed
using the Normal approximation as described in~\citep{Cram},
Chapter~34.

The experimental results are shown in dashed lines. The solid
lines correspond to the lower and upper bounds given by the
expressions~(\ref{c2}) and (\ref{t_above}). The plot shows that
upper bound~(\ref{t_above}) gives a good approximation to the
value of~$z_m^{(t)}$ even if the population size is not large. The
lower bound~(\ref{c2}) coincides with the experimental results
when~${\lambda=1}$, up to a minor sampling error.

\begin{figure}[htbr]
\begin{center}
\includegraphics[height=6.03cm, width=10.98cm]{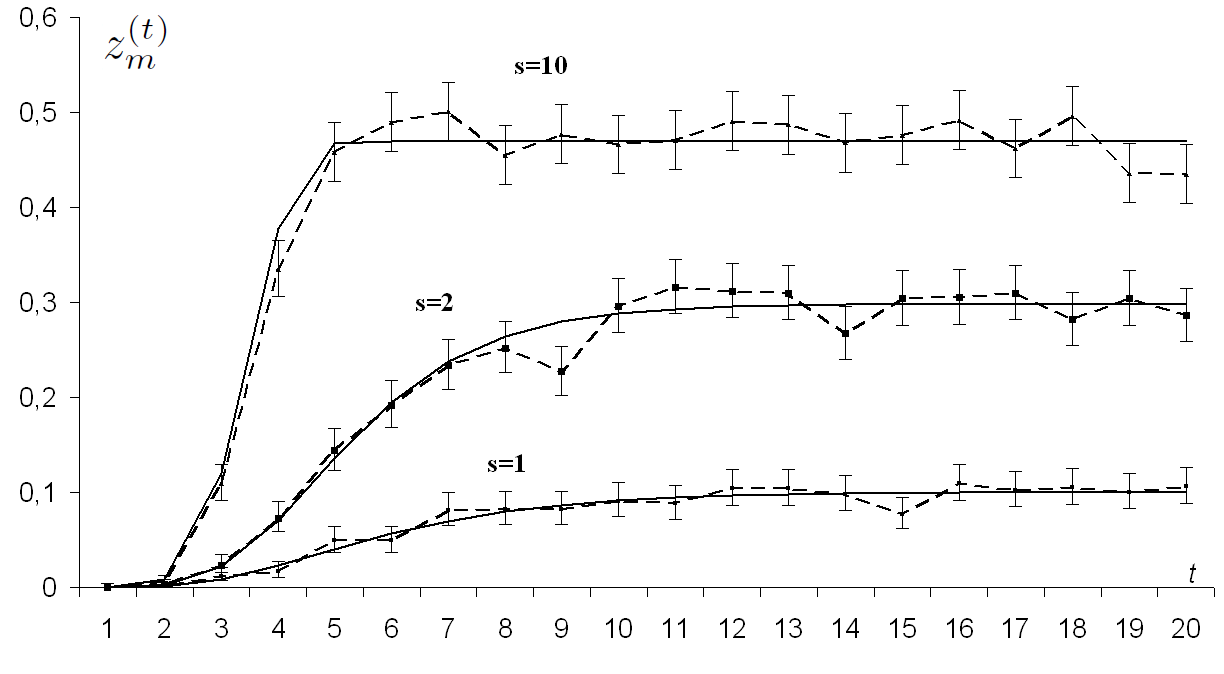}
\caption{Average proportion of optimal solutions to \vcp and the
theoretical upper bound, as functions of the iteration number.
Here $\lambda=100$, $s=1,2$ and~10.} \label{fig:different_s}
\end{center}
\end{figure}

Another series of experiments was carried out to compare the
behavior of EAs with different tournament sizes.
Figure~\ref{fig:different_s} presents the experimental results for
1000 runs of the EA with $p_{\rm m}=0.1$, $\lambda=100$ and
$\z^{(0)}={\bf 0}$ solving the \vcp on $G(8)$. This plot
demonstrates the increase in the average proportion of the optimal
genotypes as a function of the tournament size, which is
consistent with Theorem~\ref{th:tourn_size}. The 95\%-confidence
intervals are found as described above.

\subsection{Lower Bound for Randomized Local Search on Unimodal Functions.}
\label{subsec:unimod}

First of all let us describe a {\em Randomized Local Search}
algorithm~(RLS) which will be implicitly studied in this
subsection. At each iteration of RLS the current genotype~$x$ is
stored. In the beginning of RLS execution, $x$ is initialized with
some probability distribution (e.g. uniformly over~$\mathcal X$).
An iteration of RLS consists in building an offspring~$y$ of~$x$
by flipping exactly one randomly chosen bit in~$x$. If $\phi(y)
\ge \phi(x)$ then $x$ is replaced by the new genotype~$y$. The
process continues until some termination condition is met.

Below we will illustrate the usage of Theorem~\ref{th:straight} on
the class of $\ell$-\unimodal functions. In this class, each
function has exactly $\ell$ distinctive fitness values
$\phi_0<\phi_1<\dots<\phi_{\ell-1}$, and each solution in the
search space is either optimal or its fitness may be improved by
flipping a single bit. Naturally we assume that~$m=\ell-1$ and
that level~$A_m$ consists of optimal solutions.

As a mutation operator in the EA we will use a routine denoted
by~${\rm Mut}_{\rm RLS}$: given a genotype~$g$, this routine first
changes one randomly chosen gene and if this modification improves
the genotype fitness, then~${\rm Mut}_{\rm RLS}$ outputs the
modified genotype, otherwise~${\rm Mut}_{\rm RLS}(g)$ outputs the
genotype~$g$ unchanged. Note that in the case of~$\lambda=1,$ the
EA with~${\rm Mut}_{\rm RLS}$ mutation becomes a version of RLS.
The lower bounds from Section~\ref{sec:bounds} are tight
for~$\lambda=1$ (which implies~$s=1$), therefore the following
analysis in this subsection may be viewed primarily as a study of
Randomized Local Search.

Mutation operator~${\rm Mut}_{\rm RLS}$ never decreases the
genotype fitness and improves any non-optimal genotype with
probability at least~$1/n$, so we have $\alpha_{ij}=1$ for all
$i={1,\dots,m}, \ j={0,\dots,i}$ and $\alpha_{i,i+1} = 1/n$ for
$i=0,\dots,m-1$. The chances for improvements by more that one
fitness level are not foreseeable, so we put $\alpha_{ij} = 0$ for
all $i=0,\dots,m-2, \ j=i+2,\dots,m$. Note that this matrix~$\A$
is monotone.

Now $\alpha=(1/n,0,\dots,0)$ and the matrix~$\W$ consists of the
following elements:
$$
w_{ij}=\alpha_{ij}-\alpha_{i-1,j} = \cases{1/n & \mbox{\rm if }
$i=j+1$;
 \cr
1-1/n & \mbox{\rm if } $i=j$.
 \cr
0 & \mbox{\rm otherwise }.
 }
$$
In order to apply Theorem~\ref{th:straight} we also need to choose
an appropriate matrix norm and evaluate this norm for matrix~$\W$.
In this particular application we will use $||\cdot||_2,$ which is
the matrix norm induced by the Euclidean vector norm in~$\R^m$. It
is well-known that for any matrix~$\W$ holds
$||\W||_2=\sqrt{\lambda_{\max}},$ where $\lambda_{\max}$ is the
maximal eigenvalue of matrix $\W\W^T.$ Here and below $\W^T$
denotes the transpose of matrix $\W.$

It is easy to check that matrix~$\W\W^T$ is composed of zero
elements everywhere except for $m$ diagonal elements, $m-1$
superdiagonal and $m-1$ subdiagonal elements. In particular, it
has identical elements~$(1+(n-1)^2)/n^2$ on the diagonal and all
superdiagonal and subdiagonal elements are equal to~$(n-1)/n^2$.
This matrix~$\W\W^T$ belongs to the class of tridiagonal Toeplitz
matrices and its maximal eigenvalue is
$$
\lambda_{\max}=\frac{1+(n-1)^2}{n^2} + \frac{2(n-1)}{n^2} \
\cos\frac{\pi}{\ell}.
$$
(see Theorem~\ref{th:toeplitz} in the appendix). Therefore
$$
||\W||_2=\sqrt{1-\frac{2(n-1)}{n^2} \left(1-\cos
\frac{\pi}{\ell}\right)}.
$$
So $||\W||_2<1$ and since matrix~$\A$ is monotone we can apply
Theorem~\ref{th:straight}.

Let us denote ${\bf e}:=(1,1,\dots,1)\in \R^m$. The vector ${\bf
v}={\bf e}$ satisfies the equation ${\bf v}=\alpha({\bf
I}-\W)^{-1}$ and since $||\W||_2<1$, the right-hand side in
inequality~(\ref{c2}) of Theorem~\ref{th:straight} tends to~${\bf
e}$ as ${t\to \infty}$.

In order to obtain an explicit lower bound on $\E[z_m^{(t)}]$ for
any given~$t$, we will evaluate the speed of convergence of the
right-hand side in inequality~(\ref{c2}) to~{\bf e}. Note that by
properties of matrix norms we have
\begin{equation} \label{norm_bound}
||{\bf e} \W^t||_2 \le ||{\bf e}||_2 \cdot ||\W||_2^t
=\sqrt{m}~||\W||_2^t.
\end{equation}
Thus for any distribution of initial population
Theorem~\ref{th:straight} gives a lower bound
$$
\E[\z^{(t)}]\ge
 {\bf e}({\bf I}-\W^t)\ge
  {\bf e} - \sqrt{m}~||\W||_2^t \cdot {\bf e},
$$
where the last inequality holds because each component of vector
${\bf e}\W^t$ is upper-bounded by $||{\bf e} \W^t||_2$ which is at
most~$\sqrt{m}~||\W||_2^t$ by inequality~(\ref{norm_bound}).

Finally, independently of population size~$\lambda$ and tournament
size~$s$ we get a lower bound for the proportion of optimal
genotypes in the EA population:
\begin{equation}\label{eqn:lb_unimodal}
\E[z_m^{(t)}] \ge 1-\sqrt{\ell-1}\left(1-\frac{2(n-1)}{n^2} \left(
1 - \cos \frac{\pi}{\ell}\right) \right)^{t/2}.
\end{equation}
The Taylor expansion for $\cos(x)$ gives
$$
\cos\frac{\pi}{\ell} \le 1-\frac{\pi^2}{2{\ell}^2}
+\frac{\pi^4}{24{\ell}^4}\le 1- \frac{\pi^2 n^2}{2{\ell}^4}.
$$
Now since $\sqrt{1-x}\le 1-x/2$ and $\ln(1-x)\le -x,$ we obtain
$$
\E[z_m^{(t)}] \ge
 1-\sqrt{\ell-1}\left(1-\frac{\pi^2(n-1) (\ell-1)^2}{2{\ell}^4 n^2} \right)^{t}\ge
 1-\exp\left\{\frac{\ln (\ell-1)}{2} - \frac{t\pi^2}{{\ell}^2 n}
 \left(1-\frac{2}{\ell}\right)\right\}.
$$
In the case of RLS, i.e. when~$\lambda=1$, this gives the
following tail bound
\begin{corollary}\label{cor:tail_unimodal}
The probability that the maximum of a fitness function from
$\ell$-\unimodal is first reached after more than~$t$ iterations
of RLS is at most $\sqrt{e}\ ^{\ln (\ell-1) - t{\ell}^{-2}n^{-1}
(\pi^2-20{\ell}^{-1})}$.
\end{corollary}

A positive feature of this tail bound is that it approaches to~0
exponentially fast in~$t$. A weakness of
Corollary~\ref{cor:tail_unimodal} is that its bound is grater
than~1 (and therefore useless) when $t<\ln(\ell-1){\ell}^2 n
/(\pi^2-20{\ell}^{-1}).$ The obtained tail bound may be improved
for some relatively small~$t$ using the expected RLS runtime bound
and Markov inequality. Let~$T$ denote the number of fitness
evaluations made in RLS until the optimum is achieved. Then the
RLS runtime~$\E[T]\le n(\ell-1)$ since each fitness level requires
on average at most~$n$ iterations of RLS. By Markov inequality we
have $\Pr\{T\ge t\} \le n(\ell-1)/t$. This tail bound becomes
meaningful as soon as $t$ reaches $n(\ell-1)$ but it does not give
an exponential convergence and therefore yields to
Corollary~\ref{cor:tail_unimodal} for large~$t$. It would be
interesting to compare our tail bounds to those obtainable by the
approach from~\citep{LehreWitt2014} but tight analysis of RLS is
beyond the scope of this paper.

\subsection{Lower Bounds and Runtime Analysis for \twosat Problem}
\label{subsec:2SAT}

The Satisfiability problem~(SAT) in general is known to be
NP-complete~\citep{GJ}, but it is polynomially solvable in the
special case denoted by \twosat: given a Boolean formula with CNF
where each clause contains at most two literals, find out whether
a satisfying assignment of variables exists.

Let~$n$ be the number of logical variables and let $m$ be the
number of clauses in the CNF. A natural encoding of solutions is a
binary string~$g$ where $g_i=1$ if the~$i$-th logical variable has
the value~"true" and otherwise~$g_i=0.$

We consider an EA with the tournament size $s=1$ and the following
mutation operator~${\rm Mut}_{\rm SAT}$: Draw randomly a clause
which is not satisfied, choose one variable among the variables of
the clause at random, and modify this variable. Otherwise keep the
solution unchanged. This method of random perturbation was
proposed in the randomized algorithm of~\cite{Pap91} for \twosat
which has the runtime $O(n^2),$ if the CNF is satisfiable. A
generalization of the algorithm from~\citep{Pap91} to the general
case of SAT, known as WalkSat algorithm, shows competitive
experimental results~\citep{Selman95localsearch}. In the special
case of SAT, where each clause contains at most~$k$ literals,
which is denoted by $k$-SAT, algorithm WalkSat has a runtime
bound~$O((2-2/k)^k)$~\citep{Scho}.

A fitness function does not influence the EA execution when $s=1$
but it will be useful for our theoretical analysis. Let us assume
that $\phi(g)$ equals the Hamming distance to a satisfying
assignment~$g^*$. Here and below, we assume that at least one
satisfying assignment~$g^*$ exists.

For any non-satisfying truth assignment the improvement
probability is~1/2, so we can apply the following monotone bounds:
$\alpha_{ij}=1$ for all $i={1,\dots,m}, \ j={0,\dots,i-1}$;
 $ \alpha_{i,i+1} = 1/2$ for $i=0,\dots,m-1$;
$$
\alpha_{ii} = \cases{1/2 & \mbox{\rm if } $i=1,\dots,m-1$;
 \cr
1 & \mbox{\rm if } $i=m$;
 }
$$
$\alpha_{ij} = 0,$ $i=0,\dots,m-2, \ j=i+2,\dots,m$. These lower
bounds define the Markov chain transition probabilities~$\T$ with
$t_{ij}=\alpha_{ij} - \alpha_{i,j+1}, \ i=0,\dots,m, \
j=0,\dots,m-1$ and $t_{im}=\alpha_{im}, \ i=0,\dots,m$ according
to Subsection~\ref{subsec:lower_bounds}. It turns out that this
matrix~$\T$ is the same as the transition matrix of the symmetric
Gambler's Ruin random walk with one reflecting barrier (state~0)
and one absorbing barrier (state~$m$): $t_{0,1} = 1,$ $t_{i,i+1} =
t_{i,i-1}=1/2$ for $i=1,\dots,m-1$, $t_{mm}=1$, all other
elements~$t_{ij}$ are equal to zero. The result from~\citep{Pap91}
implies that, regardless of the initial state, there exists a
constant~$c>0$, such that after~$c n^2$ transitions the absorbing
probability of this random walk is at least~1/2. This means
that~$p^{(cn^2)}_m\ge 1/2$ and the $m$-th component of the
vector~$\p^{(0)} {\bf T}^t {\bf L}$ is at least~$1/2$ as well.
Therefore Theorem~\ref{thm:T} yields

\begin{corollary} \label{cor:2sat}
If the EA for \twosat has the tournament size $s=1$ and the
mutation operator~${\rm Mut}_{\rm SAT}$ then the probability to
generate a satisfying assignment in population~$X^{cn^2}$ is at
least~$1/2$ for some constant~$c>0.$
\end{corollary}

It makes sense to apply Theorem~\ref{thm:T} only in the case of
$s=1$ in this example, since for $s>1$ the tournament selection is
impossible without computing the Hamming distance to a satisfying
assignment which is unknown.

If the EA with~$s=1$ and mutation ${\rm Mut}_{\rm SAT}$ is
restarted every~$t_{\max}$ iterations and $t_{\max}=cn^2$, then
the overall runtime of this iterated~EA is~$O(\lambda n^{2})$ by
Corollary~\ref{cor:2sat} and Markov inequality. Note that
Corollary~\ref{cor:2sat} holds for any distribution of the initial
population, so the runtime bound~$O(\lambda n^{2})$ applies to the
EA without restarts as well. In a similar way the EA with ${\rm
Mut}_{\rm SAT}$ can simulate the randomized algorithm of
Sch\"{o}ning~\citep{Scho} for $k$-SAT with runtime~$O((2-2/k)^k)$.

\subsection{Lower Bounds and Runtime Analysis for Balas Set Cover Problems}
\label{subsec:Balas}

In general the set cover problem~(\scp) is formulated as follows.
Given: a ground set $M$ and a set of covering subsets
$M_j\subseteq M$, with indices $j\in U:=\{1,\ldots,n\}$. A subset
of indices $J \subseteq U$ is called a {\it cover} if $\cup_{j \in
J} {M_j}=M.$ The goal is to find a cover of minimum cardinality.
In what follows, for any $i\in M$ we denote by~$N_i$ the set of
numbers of the subsets that cover an element~$i$, \ie $N_i=\{j: i
\in M_j\}$. Note that an instance of SCP may be defined by a
family of subsets~$\{M_j\}$ or, alternatively, by a family of
subsets~$\{N_i\}$.

Suppose the {\it binary representation} of the \scp solutions is
used, i.e. genes ${g_j\in\{0,1\},} j \in U$ are the indicators of
the elements from~$U$, so that $J(g)=\{j\in U\ : \ g_j=1\}$. If
$J(g)$ is a cover then we assign its fitness $\phi(g)={n}-|J(g)|$;
otherwise $\phi(g)=r(g)$, where $r(g)<0$ is a decreasing function
of the number of non-covered elements from~$M.$

Consider a family ${\mathcal B}(n,k)$ of set cover problems
introduced by~\cite{Ba84}. Here it is assumed that~$M=\{1,\dots,{n
\choose n-k+1}\}$ and that all $(n-k+1)$-element subsets of~$U$
are given as subsets $N_1,N_2,...,N_{|M|}$. Thus any collection of
less than $k$ elements from~$U$ belongs to $U\backslash N_i$ for
some~$i\in M$ and does not cover the element~${i\in M}$. At the
same time any subset $J \subseteq U$ of size~$k$ covers all
elements of~$M$ and therefore it is an optimal cover. Larger
subsets are non-optimal covers.

Since any $k$-element subset of~$U$ is an optimal cover, family
${\mathcal B}(n,k)$ is solvable trivially. Nevertheless this
family is known to be hard for general-purpose integer programming
algorithms~\citep{Ba84,Saiko}. In particular, it was shown
in~\citep{Saiko} that problems from this class are hard to solve
using the $L$-class enumeration method~\citep{Kolo}. When~$n$ is
even and~$k=n/2$, the $L$-class enumeration method needs an
exponential number of iterations in~$n$. In what follows we
analyze the EA in this special case.

Note that any $i$-element subset~$J\subseteq U$ for $i<k$ leaves
${n-i \choose n-(k-1)}$ elements of the ground set uncovered,
regardless of the choice of elements in~$J$. So in the case of
tournament selection, equivalently to studying the EA on
family~${\mathcal B}({n},{n/2})$ we may study the EA where the
fitness is given by a function of unitation, so that
$$  \phi(g) = \left\{
\begin{array}{ll}
 R(||g||_1) &  \mbox{\rm if } \ \ ||g||_1 \ge n/2;\\
 L(||g||_1) &
\mbox{\rm otherwise,}
\end{array} \right.
$$
where function $R$ is decreasing, function $L$ is increasing and
$L(\frac{n}{2}-1)<R(n).$

Consider the point mutation operator with tunable parameter~$q>0$
defined in Subsection~\ref{subsec:lower_bounds}. Let $m={n/2}$ and
let the thresholds $\phi_0, \phi_1,...,\phi_{m}$ be equal to
fitness of genotypes that contain~$0,1,...,m$ genes~"1"
accordingly. Note that $J(g)$ is a cover iff $\phi(g) \ge \phi_m$.

We have the following lower bounds: $\alpha_{ij}=1$ for all
$i={1,\dots,m}, \ j={0,\dots,i-1}$;
 $ \alpha_{i,i+1} = (1-q)(n-i)/n$ for $i=0,\dots,m-1$;
$$
\alpha_{ii} = \cases{
 q+ \alpha_{i,i+1} & \mbox{\rm if }
$i=1,\dots,m-1$;
 \cr
q & \mbox{\rm if } $i=m$;
 }
$$
$\alpha_{ij} = 0,$ $i=0,\dots,m-2, \ j=i+2,\dots,m$. These lower
bounds $\alpha_{ij}$ coincide with the corresponding cumulative
transition probabilities except for level~$i=m$, where we
pessimistically assume~$\alpha_{mm}=q$ (in fact we could safely
put $\alpha_{mm}=0.5(1-q)+q$ but $\alpha_{mm}=q$ is chosen to
match the model of Ehrenfests in what follows). It is easy to
verify that~$\A$ satisfies the monotonicity condition when ${q}\ge
1/(n+1)$ just as we verified this in the example of monotone
mutation in Subsection~\ref{subsec:lower_bounds}.

In case we are interested in runtime bounds for the EA, rather
than expected values of vector~$\z^{(t)}$, we can
assume~$\alpha'_{mm}=1$. All other non-zero lower
bounds~$\alpha_{ij}$ defined above could be relaxed by
putting~$\alpha'_{ij}=1/2$. In this case we would have the
associated Markov chain with a transition matrix~$\T',$ the same
as in Subsection~\ref{subsec:2SAT}, resulting in the same EA
runtime bound~$O(\lambda n^2)$. We shall avoid these
simplifications, however, in order to obtain a tighter runtime
bound by means of the following corollary.

\begin{corollary} \label{cor:Balas}
Suppose that the EA with a tournament size~$s\ge 1$ uses the point
mutation operator with parameter~${q\ge 1/(n+1)}$. Then given
$X^0=({\bf 0},\dots,{\bf 0}),$ there exists a constant~$c$, such
that the probability to reach an optimum of problem~${\mathcal
B}(n,n/2)$ within $\lceil c n \ln n \rceil$ iterations
is~$\Omega(n^{-0.5}).$
\end{corollary}

To prove this corollary, first we will  obtain a lower bound
on~$\E[z_m^{(t)}]$ for~$t\to\infty$, using Theorem~\ref{thm:T} and
the stationary distribution of the associated Markov chain ${\bf
p}^{(t)} = {\bf p}^{(0)} \ {\bf T}^t.$ After that, analogously to
the proof of Corollary~\ref{cor:tail_unimodal}, we will compute a
lower bound on~$\E[z_m^{(t)}]$ for finite~$t$, using
Theorem~\ref{th:straight}.

{\bf Proof of Corollary~\ref{cor:Balas}.}  The Markov chain
associated to the set of lower bounds~$\alpha_{ij}$ defined above
has the following nonzero transition probabilities
$$
t_{ii}=q, \ \ t_{i,i-1} = (1-q)i/n, \ \ t_{i,i+1}=(1-q)(1-i/n), \
\ i=1,\dots,m-1,
$$
$$
t_{0,1} = 1-q, \ \ t_{m,m-1} =1-q, \ \ t_{mm}=q.
$$
All other elements of matrix~$\T$ are equal to zero.

The stationary distribution of the associated Markov chain may be
found from the well-known model for diffusion of P.~Ehrenfest and
T.~Ehrenfest. Consider $n$ molecules in a rectangular container
divided into two equal parts A and B. At any time~$t$, one
randomly chosen molecule moves to another part. The state of the
system is defined by the number of molecules~$j, \ j=0,\dots,n,$
in container~A. The corresponding random walk has transition
probabilities
$$
\tau_{j,j-1} = j/n, \ \ \tau_{j,j+1}=1-j/n, \ \ j=1,\dots,n-1,
$$
$$
\tau_{0,1} = 1, \ \ \tau_{n,n-1} =1.
$$
The stationary distribution in Ehrenfests model (see
e.g~\citep{Feller}, chapter.~15, \S~6) is given by $\pi_j:={{n
\choose j}}/{2^{n}}, \ j=0,\dots,n.$ Grouping each couple of
symmetric states (i.e. the state where A contains~$j$ molecules, B
contains~$n-j$ molecules and the state where A contains~$n-j$
molecules and B contains~$j$ molecules, $j=0,\dots,n/2$) into one
state we conclude that the Markov chain with transition
matrix~$\T$ has the stationary distribution~${\bf
u}=(2\pi_1,\dots,2\pi_m)$ for any $q<1$. So by
Theorem~\ref{thm:T}, vector~${\bf u}{\bf L}$ is the limiting lower
bound for $\E[\z^{(t)}]$ as $t\to \infty$.

We are interested in transient behavior of the EA, so we will
obtain a lower bound for the expected population
vector~$\E[\z^{(t)}],$ given a finite~$t$, using
Theorem~\ref{th:straight}. Consider the matrix norm~$||{\bf
W}||_{\infty}=\max_{i=1,\dots,m} \sum_{j=1}^m |w_{ij}|$ which is
associated to the vector norm $||\cdot||_1$ in the case of
left-hand side multiplication of matrices by vectors. For the
matrix~${\bf W}$, corresponding to the set of lower
bounds~$\alpha_{ij},$ defined above, we have~$||{\bf
W}||_{\infty}=1-2(1-{q})/n$, \ie the condition $\lim\limits_{t \to
\infty} ||\W^t||_{\infty}=0$ is satisfied for any~${q}<1$.

Let us find the vector ${\bf v}={\alpha}({\bf I}-{\bf W})^{-1}$,
which is the limit of the right-hand side in inequality~(\ref{c2})
as $t \to \infty$. To this end, it suffices to solve the system of
equations
\begin{equation}\label{stationary1}
-v_{i-1} \frac{n-i+1}{n} + v_i \frac{n+1}{n} - v_{i+1} \frac{i}{n}
=0, \ \ i=2,\dots,\frac{n}{2}-1,
\end{equation}
\begin{equation}\label{stationary2}
 v_{1} \frac{n+1}{n} - v_{2} \frac{1}{n} =1, \ \ \ \ \
 -v_{n/2-1} \frac{n+2}{2n} + v_{n/2} \frac{n-1}{n} =0.
\end{equation}
Recall that the right-hand sides in inequalities~(\ref{c2}) and
(\ref{eq:fromMarkov}) of Theorems~\ref{th:straight}
and~\ref{thm:T} are equal, given equal matrices~$\A$. This
suggests to put ${\bf v}={\bf u} {\bf L}$, \ie
\begin{equation} \label{Ehrenfest}
v_i=\sum_{\ell=i}^{n/2} {n \choose {\ell}} \frac{1}{2^{n-1}}, \ \
i=1,\dots,\frac{n}{2}.
\end{equation}

Again let ${\bf e}=(1,\dots,1)$. By properties of the norms under
consideration, ${\bf v W}^t \le ||{\bf v W}^t||_1 {\bf e} \le
||{\bf v}||_1 \cdot ||{\bf W}||^t_{\infty} {\bf e} \le m ||{\bf
W}||^t_{\infty} {\bf e}$, so by Theorem~\ref{th:straight}
$$
\E[{\z}^{(t)}]\geq \E[{\z}^{(0)}] {\bf W}^t+{\alpha}({\bf I}-{\bf
W})^{-1}({\bf I}-{\bf W}^t)
 \geq {\alpha}({\bf I}-{\bf W})^{-1}-{\alpha}({\bf I}-{\bf
W})^{-1} \W^t
 \geq {\bf v} - m ||{\bf
W}||_{\infty}^t {\bf e}
$$
for any~$t$. With ${q}=1/(n+1),$ the average proportion of
feasible genotypes is lower-bounded by $v_m -
m\left(\frac{n-1}{n+1}\right)^t$ since
$||\W||_{\infty}=1-\frac{2(1-q)}{n}=\frac{n-1}{n+1}$.
Using~(\ref{Ehrenfest}) and Stirling's inequality
$\sqrt{2\pi}n^{n+0.5} e^{-n} \le n! \le e n^{n+0.5} e^{-n}$ we
conclude that $v_m = \frac{{n \choose n/2}}{2^{n-1}}
=\Omega(n^{-0.5})$. Now assuming that a constant~$c$ is so lagre
that $cn\ln n \ge \frac{n+1}{2} \ln \frac{n}{v_m}$, for $t=\lceil
c n \ln n \rceil$ we have
$$
\frac{v_m}{n} \ge
 \left(\frac{1}{e}\right)^{\frac{2t}{n+1}} \ge
 \left(\left(1-\frac{2}{n+1}\right)^{\frac{n+1}{2}}\right)^{\frac{2t}{n+1}}=\left(\frac{n-1}{n+1}\right)^t,
$$
so $\frac{n}{2}\left(\frac{n-1}{n+1}\right)^t \le \frac{v_m}{2}$
and $\E[{z}^{(t)}_m]\geq \frac{v_m}{2}$.

By assumption the initial population consists of all-zero strings.
Therefore the presence of at least one individual from~$H_m$ in
the current population implies that an optimal solution to a
problem~${\mathcal B}(n,n/2)$ was already found at least once.
Thus, in view of Proposition~\ref{vose}, after~$\lceil c n \ln n
\rceil$ iterations of the EA, the probability of finding an
optimum is
at least~$\Omega(n^{-0.5})$ and the corollary is proved. $\Box$\\

If the EA is restarted with $X^0=({\bf 0},\dots,{\bf 0})$ every
$t_{\max}=\lceil c n \ln n \rceil$ iterations, then by Markov
inequality the overall runtime of this iterated~EA is $O(\lambda
n^{1.5} \log n)$ for any~$\lambda$.

The tools for the non-elitist EA analysis
from~\citep{bib:cdel14,bib:dl16,er2016_proc} can be adjusted to
upper-bound the runtime of the EA on~${\mathcal B}(n,n/2),$ but in
such a case, a non-zero selection pressure would be required with
a sufficiently large~$s$ and the results would hold only
for~$\lambda=\Omega(\log n)$.

\subsection{Upper Bound on Proportion of Optimal Genotypes in Case of \onemax}
\label{subsec:onemax_tail}

The upper bounds on vector~$\z^{(t)}$ obtained in
Proposition~\ref{prop:upper_bound} are not likely to be suitable
for obtaining the  lower bounds on runtime of the EA in absolute
terms due to nonlinearity in the right-hand side
of~(\ref{t_above}). There are other methods for finding such lower
bounds on the runtime proposed e.g. in
~\citep{Badkobeh2014,Lehre2010PopNegDrift,Sudholt13}. The upper
bounds on vector~$\z^{(t)}$ however may be used for comparison of
the EA to the \onelambdaEA and the \oneplusoneEA as it was
suggested in Proposition~\ref{prop:upper_bound1}.

To illustrate such a comparison let us consider the EA with
bitwise mutation operator~${\rm Mut}$ in the case of \onemax
fitness function and assume that~${\phi_i:=i,} \ {i=0,\dots,n.}$
Analogously to the notation form Section~\ref{sec:bounds},
$P^{(\tau)}_n$ and $Q^{(\tau)}_n$ will stand for the probability
to have an optimal current individual on iteration~$\tau$ of
\onelambdaEA and on iteration~$\tau$ of the \oneplusoneEA,
respectively. In these algorithms we assume that the bitwise
mutation operator~${\rm Mut}'={\rm Mut}$ is used
and the initial solution is chosen uniformly from~$\mathcal X$.
Proposition~\ref{prop:upper_bound1} yields the following

\begin{corollary} \label{cor:upper_bound_OM}
Suppose that  the fitness function is \onemax and the initial
population of the EA consists of~$\lambda$ copies of the same
solution, chosen uniformly from~$\mathcal X$, and the EA uses the
bitwise mutation operator with~$p_{\rm m}=1/n.$ Then for any $t\ge
0$ holds
$$
\E[z_n^{(t+1)}]\le
\frac{1}{e} - \frac{n-2}{e(n-1)} (1-P^{(t)}_n)^s
 \le \frac{1}{e} -
\frac{n-2}{e(n-1)} (1-Q^{(t\lambda)}_n)^s.
$$
In particular, if the tournament size $s=2$ then
$\E[z_n^{(t+1)}]\le {0.74 P^{(t)}_n + O(n^{-1})}$ and
$\E[z_n^{(t+1)}]\le {0.74 Q^{(t\lambda)}_n + O(n^{-1})}.$
\end{corollary}

{\bf Proof.} In the case of \onemax fitness function the bitwise
mutation operator with~$p_{\rm m}=1/n$ is monotone~\citep{BE08}.
Application of Proposition~\ref{prop:upper_bound1} yields
$\E[z_n^{(t+1)}]\le {\gamma_{nn} - (\gamma_{nn}-\gamma_{n-1,n})
(1-P^{(t)}_n)^s}$ for the cumulative transition
probabilities~$\gamma_{ij}$ associated with this monotone mutation
operator. It is easy to see that $\gamma_{n-1,n} \le e^{-1}/(n-1)$
and $\gamma_{n,n} \le e^{-1},$ since $(1-1/n)^n\le e^{-1}.$ Thus,
for the \onelambdaEA
$$
\E[z_n^{(t+1)}]\le \frac{1}{e} -
\left(\frac{1}{e}-\frac{1}{e(n-1)}\right) (1-P^{(t)}_n)^s
$$
as required. In the case of $s=2$ this inequality implies that
$\E[z_n^{(t+1)}] \le \frac{2(n-2)}{e(n-1)} P_n^{(t)} +
\frac{1}{e(n-1)} \le 0.74 P^{(t)}_n + O(n^{-1}).$ The result for
\oneplusoneEA follows analogously.
$\Box$\\

A superiority of the \oneplusoneEA over other evolutionary
algorithms in the case of \onemax fitness function and bitwise
mutation with~$p_{\rm m}\le 0.5$ is well-known
from~\citep{bor01,BE08,Sudholt13}.
Corollary~\ref{cor:upper_bound_OM} allows to measure the
superiority of \oneplusoneEA and the \onelambdaEA over the EA in
terms of tail bounds. Note that the tail bounds for the
\oneplusoneEA on \onemax are well studied. In particular, the tail
bound from~\citep{LehreWitt2014} implies that there exists such
constant~$c>0$ that for any~$r\ge 0$ and $\tau<e n \ln n - cn - r
e n$ holds $Q^{(\tau)}_n\le e^{-r/2}$.

\section{Conclusions}
\label{sec:conc}

In this paper, we presented an approximating model of non-elitist
mutation-based EA with tournament selection and obtained upper and
lower bounds on proportion of sufficiently good genotypes in
population using this model. In the special case of monotone
mutation operator, the obtained bounds become tight in different
situations. The analysis of infinite population EA with monotone
mutation suggests an optimal selection mechanism that actually
converts the EA into the \onelambdaEA.

Applications of the obtained general lower bounds give an
exponentially vanishing tail bound for the Randomized Local Search
on unimodal functions and new runtime bounds for the EAs on the
2-satisfiability problem and on a family of set covering problems
proposed by E.~Balas.

It is expected that the further research will involve applications
of the proposed approach to other combinatorial optimization
problems, in particular, the problems with regular structure.

Most of the lower and upper bounds on expected proportions of
genotypes, obtained in this paper, do not take the tournament size
into account. It remains an open research question of how to
construct the tighter bounds w.r.t. the tournament size.
The subsequent research might benefit from joining the analysis of
expectation of population vector with some variance analysis.

It is of interest to compare the tail bounds established in
Subsections~\ref{subsec:unimod} and~\ref{subsec:onemax_tail} to
the tail bounds obtainable using other techniques,
e.g.~\citep{LehreWitt2014}.

Another open question is how to incorporate the crossover operator
into the approximating model. For some types of crossover
operators, such as those based on solving the optimal
recombination problem~\citep{ErKov14}, the lower bounds from this
paper may be easily extended, ignoring the improving capacity of
crossover. It is important, however, to take the positive effect
of crossover into account and it is not clear how the monotonicity
conditions could be meaningfully extended for this purpose.

\section*{Appendix.}

In this appendix, we reproduce two results from~\citep{BE01}
and~\citep{bor01} which are used in Section~\ref{sec:bounds} and a
well-known result on eigenvalues of thridiagonal Toeplitz
matrices.

The algorithms \onelambdaEA and \oneplusoneEA and probabilities
$P_j^{(\tau)}$ and $Q_j^{(\tau)}$, $j=1,\dots,m,$ $\tau=0,1,\dots$
are defined as in Section~\ref{sec:bounds}. For the \onelambdaEA
and for the \oneplusoneEA we also define the vectors of
probabilities: $
\PP^{(\tau)}=\left(P_1^{(\tau)},...,P_m^{(\tau)}\right),
$
 $ \Q^{(\tau)}=\left(Q_1^{(\tau)},...,Q_m^{(\tau)}\right).
$

The following Theorem~\ref{EScompare} from~\citep{BE01} shows a
superiority of the \oneplusoneEA over the \onelambdaEA in the case
of monotone mutation operator. For a fair comparison of the
algorithms \onelambdaEA and \oneplusoneEA here we allow both of
them to make the same number of evaluations of the fitness
function, equal to~$t\lambda$.

\begin{theorem}
\label{EScompare} Suppose that the same monotone mutation
operator~${\rm Mut}'$ is used in the \oneplusoneEA and in the
\onelambdaEA and ${\Q^{(0)}\geq \PP^{(0)} }.$ Then
$\Q^{(t\lambda)}\geq \PP^{(t)}$ for any $t\ge 0$.
\end{theorem}


The following theorem from~\citep{bor01} compares the distribution
of a fittest individual~$g_*^{(t)}$ in the EA population~$t$ over
Lebesgue subsets compares to such a distribution of the
\onelambdaEA. Let us define a vector~${\bf R}^{(t)}$ for the EA,
analogously to vectors $\PP^{(t)}$ and $\Q^{(t)}$:
$$
{\bf R}^{(t)}:=\left(\Pr\{g_*^{(t)}\in
H_1\},\dots,\Pr\{g_*^{(t)}\in H_m\}\right).
$$

\begin{theorem} \label{th:bor}
Suppose that the EA and the \onelambdaEA use the same monotone
mutation operator~${\rm Mut}$ and ${\bf R}^{(0)}\le \PP^{(0)}$.
Then for any~$t\ge 0$ holds ${\bf R}^{(t)}\le \PP^{(t)},$
regardless of selection operator used in the EA.
\end{theorem}

The original manuscript~\citep{bor01} is hardly accessible,
therefore we provide the proof of Theorem~\ref{th:bor} below.

{\bf Proof.} It is sufficient to consider the case of $t=1$, since
the statement for the general case will follow by induction on
$t$. Let~$b^{(1,k)}$ denote a genotype with the highest fitness
among the first~$k$ offspring of~$b^{(0)}$ and let~$g^{(1,k)}$ be
a genotype with the highest fitness
among~$g_1^{(1)},\dots,g_1^{(k)}$ in the EA population~$X^1$, for
any $k=1,\dots,\lambda$.

a) Let us first assume that $b^{(0)}\in A_i$ and $g_*^{(0)}\in
A_i$ for some fixed $i$ and let a genotype~$g'$ be chosen by the
selection operator of the EA. Then for arbitrary $j=1\dots m,$ in
view of Proposition~\ref{prop:monotone} we have:
\begin{equation}
\label{eq:9} \Pr\{{\rm Mut}(g')\not\in H_j | g_*^{(0)}\in A_i\}
\ge \Pr\{{\rm Mut}(b^{(0)})\not\in H_j | b^{(0)}\in A_i\}.
\end{equation}
Note that $\Pr\{g_*^{(1)}\not \in H_j | g_*^{(0)}\in A_i\} \ge
\Pr\{b^{(1)} \not \in H_j | b^{(0)}\in A_i\}$, which may be
established by induction on~$k=1,\dots,\lambda-1$ using the
inequality
$$
\Pr\{g^{(1,k+1)}\not \in H_j | g_*^{(0)}\in A_i\}=
 \Pr\{g^{(1,k)}\not \in H_j | g_*^{(0)}\in A_i\}
 \Pr\{{\rm Mut}(g')\not\in H_j | g_*^{(0)}\in A_i\}
$$
$$
\ge
 \Pr\{b^{(1,k)}\not \in H_j | b^{(0)}\in A_i\}
 \Pr\{{\rm Mut}(b^{(0)})\not\in H_j | b^{(0)}\in A_i\}=
\Pr\{b^{(1,k+1)}\not \in H_j | b^{(0)}\in A_i\}.
$$

b) Let us prove that $\PP^{(1)} \ge {\bf R}^{(1)}$ for arbitrary
initial distributions of the \onelambdaEA and the EA, assuming
$\PP^{(0)} = {\bf R}^{(0)}$. We use the total probability formula
and the conclusion of case a):
\[\Pr\{g_*^{(1)}\not\in H_j \}=
\sum\limits_{i=0}^m \Pr\{g_*^{(1)}\not\in H_j | g_*^{(0)}\in A_i
\}\Pr\{g_*^{(0)} \in A_i\}
\]
\vspace{-1em}
\begin{equation}
\label{equiv} \geq \sum\limits_{i=0}^m \Pr\{ b^{(1)}\not\in H_j |
 b^{(0)}\in A_i \}\Pr\{b^{(0)} \in A_i\} =
\Pr\{ b^{(1)}\not\in H_j \}.
\end{equation}

c) In general, when  $\PP^{(0)}\geq {\bf R}^{(0)}$ let us note
that according to Proposition~1 from~\citep{BE01}, in the case of
monotone mutation for any~$t\ge 1$ we can consider $\PP^{(t)}$ as
the following function on vector $\PP^{(t-1)}$:
\begin{equation}
\label{BoundP}
 P^{(t)}_j= 1-(1-\gamma_{0j})^\lambda+ \sum\limits_{i=1}^m
((1-\gamma_{i-1,j})^\lambda-(1-\gamma_{ij})^\lambda) P^{(t-1)}_i,
\ j=1,\dots,m,
\end{equation}
where~$\gamma_{ij}$ are the cumulative transition probabilities of
mutation operator~${\rm Mut}$. We denote the
relationship~(\ref{BoundP}) by $\PP^{(t)}=F(\PP^{(t-1)})$ for
brevity. Then due to nonnegativity of the multipliers of
probabilities $P_1^{(t-1)},...,P_m^{(t-1)}$ in (\ref{BoundP}), we
conclude that $\PP^{(1)}=F(\PP^{(0)}) \ge F({\bf R}^{(0)})$.
Finally note that the result of case~b) may be written as $F({\bf
R}^{(0)}) \ge {\bf R}^{(1)}$, therefore $\PP^{(1)}\ge {\bf
R}^{(1)}.$
$\Box$\\

The following result on eigenvalues of thridiagonal Toeplitz
matrices may be found e.g. in~\citep{NPR13}.

\begin{theorem} \label{th:toeplitz}
Suppose an $(n\times n)$-matrix~$\T$ is composed of zero elements
everywhere except for the diagonal elements, which equal~$\delta$,
the superdiagonal elements which equal~$\tau$ and subdiagonal
elements which equal~$\sigma$. Then all of eigenvalues of~$\T$ are
given by
$$
\lambda_{h}=\delta + 2\sqrt{\sigma \tau} \ \cos\frac{h \pi}{n+1},
\ \  h=1,\dots,n.
$$
\end{theorem}

\subsection*{Acknowledgements}

The research presented in Section~\ref{sec:app} was supported by
Russian Foundation for Basic Research grants~15-01-00785 and
16-01-00740. The author is grateful to Sergey A. Klokov, Boris A.
Rogozin and anonymous referees for helpful comments on earlier
versions of this work.

\small

\bibliographystyle{apalike}
\bibliography{paper50}

\end{document}